\documentclass[10pt,journal,compsoc]{IEEEtran}

\ifCLASSOPTIONcompsoc
  \usepackage[nocompress]{cite}
\else
  \usepackage{cite}
\fi
\ifCLASSINFOpdf
  \usepackage[pdftex]{graphicx}
  \graphicspath{{../Figures/}}
  \DeclareGraphicsExtensions{.pdf,.jpeg,.png}
\else
  \usepackage[dvips]{graphicx}
  \graphicspath{{../Figures/}}
  \DeclareGraphicsExtensions{.eps}
\fi
\usepackage{amsmath}
\usepackage{algorithmic}
\usepackage{array}
\ifCLASSOPTIONcompsoc
  \usepackage[caption=false,font=footnotesize,labelfont=sf,textfont=sf]{subfig}
\else
  \usepackage[caption=false,font=footnotesize]{subfig}
\fi
\usepackage{stfloats}
\ifCLASSOPTIONcaptionsoff
  \usepackage[nomarkers]{endfloat}
 \let\MYoriglatexcaption\caption
 \renewcommand{\caption}[2][\relax]{\MYoriglatexcaption[#2]{#2}}
\fi
\usepackage{url}
\usepackage{algorithm}
\usepackage{algorithmic}
\usepackage{booktabs}

\usepackage{xcolor}
\usepackage{amsmath}
\usepackage{enumitem}
\usepackage{csquotes}
\usepackage[export]{adjustbox}
\usepackage[caption=false]{subfig}
\usepackage{amsfonts}
\usepackage{ulem}

\begin{document}

\title{STTAR: Surgical Tool Tracking using off-the-shelf Augmented Reality Head-Mounted Displays}

\author{Alejandro~Martin-Gomez$^\dagger$,
~Haowei~Li$^\mathsection$,~Tianyu~Song,~Sheng~Yang,~Guangzhi~Wang,

~Hui~Ding,~Nassir~Navab,~\IEEEmembership{Fellow,~IEEE},%
~Zhe~Zhao,~Mehran~Armand
\IEEEcompsocitemizethanks{
\IEEEcompsocthanksitem A. Martin-Gomez$^\dagger$ and H. Li$^\mathsection$ have contributed equally to this work.\protect\\
E-mail: $^\dagger$alejandro.martin@jhu.edu, $^\mathsection$lihaowei19991202@gmail.com
\IEEEcompsocthanksitem A. Martin-Gomez, N. Navab, and M. Armand are with the Laboratory for Computational Sensing and Robotics, Whiting School of Engineering, Johns Hopkins University, United States of America.
\IEEEcompsocthanksitem M. Armand is also with the Department of Orthopaedic Surgery, Johns Hopkins University School of Medicine, United States of America.
\IEEEcompsocthanksitem H. Li, S. Yang, G. Wang, and H. Ding are with the Department of Biomedical Engineering, Tsinghua University, China.
\IEEEcompsocthanksitem Z. Zhao is with the Department of Orthopaedics, Beijing Tsinghua Changgung Hospital. School of Clinical Medicine, Tsinghua University.
\IEEEcompsocthanksitem T. Song and N. Navab are with the Chair for Computer Aided Medical Procedures and Augmented Reality, Department of Informatics, Technical University of Munich, Germany.
}
}


\IEEEtitleabstractindextext{%
\begin{abstract}
The use of Augmented Reality (AR) for navigation purposes has shown beneficial in assisting physicians during the performance of surgical procedures.
These applications commonly require knowing the pose of surgical tools and patients to provide visual information that surgeons can use during the performance of the task.
Existing medical-grade tracking systems use infrared cameras placed inside the Operating Room (OR) to identify retro-reflective markers attached to objects of interest and compute their pose.
Some commercially available AR Head-Mounted Displays (HMDs) use similar cameras for self-localization, hand tracking, and estimating the objects’ depth.
This work presents a framework that uses the built-in cameras of AR HMDs to enable accurate tracking of retro-reflective markers, such as those used in surgical procedures, without the need to integrate any additional components.
This framework is also capable of simultaneously tracking multiple tools. Our results show that the tracking and detection of the markers can be achieved with an accuracy of $0.09\pm 0.06\ mm$ on lateral translation, $0.42 \pm 0.32\ mm$ on longitudinal translation and $0.80 \pm 0.39^\circ$ for rotations around the vertical axis.
Furthermore, to showcase the relevance of the proposed framework, we evaluate the system’s performance in the context of surgical procedures.
This use case was designed to replicate the scenarios of k-wire insertions in orthopedic procedures. For evaluation, two surgeons and one biomedical researcher were provided with visual navigation, and each performed 21 injections.
Results from this use case provide comparable accuracy to those reported in the literature for AR-based navigation procedures.
\end{abstract}

}

\maketitle

\IEEEdisplaynontitleabstractindextext

%

\IEEEpeerreviewmaketitle

\IEEEraisesectionheading{\section{Introduction}\label{sec:Introduction}}

\begin{figure*}[!t]
    \centering
        \subfloat[\label{fig:Teaser_A}]{\includegraphics[page=1,width=0.65\columnwidth]{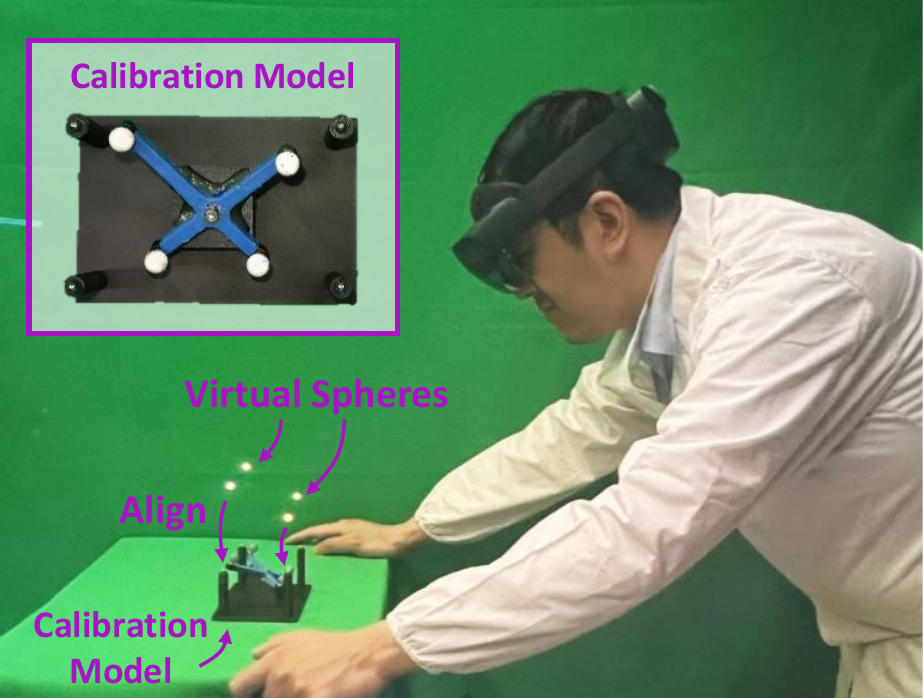}}\hfill
        \subfloat[\label{fig:Teaser_B}]{\includegraphics[page=2,width=0.65\columnwidth]{Figures/Teaser_v3.pdf}}\hfill
        \subfloat[\label{fig:Teaser_C}]{\includegraphics[page=3,width=0.65\columnwidth]{Figures/Teaser_v3.pdf}}
    \caption{\protect\subref{fig:Teaser_A} Calibration model used to align a set of virtual and real spheres. This model enables the registration of the observer's view and the image sensors. \protect\subref{fig:Teaser_B} Our tracking system enables the localization of surgical tools and anatomical models by attaching retro-reflective passive markers without integrating any other external trackers or electronics. \protect\subref{fig:Teaser_C} A surgeon can then use this method to insert surgical k-wires assisted by virtual trajectories displayed using an Augmented Reality Head-Mounted Display.}
    \label{fig:Teaser}
\end{figure*}

\IEEEPARstart{I}{n} recent years, an increasing number of Augmented Reality (AR) applications have found their use in medical domains, including educational, training, and surgical interventions \cite{kamphuis2014augmented,peters2018mixed,campisi2020augmented,Mehrfard:2021:Virtual}.
Among the multiple technologies capable of delivering augmented content, using AR Head-Mounted Displays (HMDs) has proven beneficial to aid physicians during surgical interventions \cite{fida2018augmented,deib2018image,jud2020applicability,casari2021augmented}.
The use of HMDs in the Operating Room (OR) allows for the observation of relevant content in-situ, enables delivering visual guidance and navigation, promotes the visualization and understanding of diverse bi-dimensional and three-dimensional medical imaging modalities, and facilitates training and communication \cite{rahman2020head,fotouhi2020development}.

A fundamental requirement to provide visual guidance and surgical navigation using AR is to know the three-dimensional pose of the elements involved in the task (i.e., their position and orientation).
Although multiple technologies can be used to estimate the pose of the objects of interest, optical tracking is one of the most common due to its computational simplicity, cost-effectiveness, and the availability of cameras used in AR applications \cite{koulieris2019near}.
This tracking technology uses computer vision algorithms to identify textural patterns or geometrical properties of objects of interest, from which their dimensions are well known, to estimate their pose.

In the medical context, providing surgical navigation requires knowing the pose of the patients, surgical tools, and, occasionally, the surgeons and the medical team's head.
Commercially available medical-grade tracking systems estimate the pose of the objects of interest using infrared cameras that detect the location of retro-reflective markers rigidly attached to them.
These retro-reflective markers are arranged in unique configurations to distinguish between the multiple objects observed in the scene.
Commercially available HMDs, use built-in infrared cameras for self-localization, hand tracking, and estimation of the objects' depth in the environment.
Thus, these cameras can also be used to locate the retro-reflective markers used in medical-grade tracking systems\footnote{e.g., Polaris NDI, Northern Digital Incorporated. Ontario, Canada.} \cite{Kunz:2020:IR_HoloLens_Neuro,Gsaxner:2021:IO_HoloLend_Navigation}.
While combining the accuracy provided by medical-grade tracking systems with the visualization capabilities of AR HMDs could support the integration of these devices into surgical scenarios, simultaneously maintaining the fiducial markers in the line of sight of the HMD and the tracking system in a cluttered workspace may hinder its use in the OR. 
As an alternative, using the HMD for combined visualization and tracking of these markers may mitigate such challenges.
However, the accuracy and feasibility of using cameras and sensors that may not have the accuracy and field of view of conventional tracking systems require detailed investigation.
Therefore, it is essential to establish a framework capable of performing this task and investigate the accuracy of combined tracking and visualization when only using HMDs.

This work introduces a method that uses the built-in Time-of-Flight (ToF) sensor of commercially available HMDs\footnote{\label{HoloLens}i.e., Microsoft HoloLens 2. Microsoft. Washington, USA.} to define, detect and localize passive retro-reflective markers commonly used during navigation-assisted medical procedures.
Compared to existing works, our method can track multiple objects in real-time and does not require the integration of additional hardware such as infrared LEDs to illuminate the retro-reflective markers.
In addition, we investigate the capabilities of the built-in cameras of the HoloLens2 to provide valuable insights into their strengths and limitations.
Furthermore, we present a use case in the context of medical applications to provide navigation capabilities and aid physicians performing percutaneous surgical procedures assisted by this type of HMD.
We expect these experiments can aid researchers in deciding if such devices are suitable for specific applications.\footnote{The code for our work can be accessed through the following link: \url{https://github.com/lihaowei1999/IRToolTrackingWithHololens2.git}}
\section{Related Work}\label{sec:RelatedWork}

Using commercially available AR HMDs for medical applications has found application in several surgical fields.
In spine surgery, providing surgical navigation for pedicle screw placement using AR HMDs has shown shorter placement time when compared to traditional procedures \cite{yoon2017technical} and comparable accuracy to robotic-assisted computer-navigated approaches \cite{molina2019augmented}.
In addition, it has supported spinal pedicle guide placement for cannulation without using fluoroscopy \cite{gibby2019head} and enabled the registration between the patient's anatomy and three-dimensional models containing surgical planning data \cite{liebmann2019pedicle}.
This technology has also been used to provide visual guidance during the performance of osteotomies, contributing to an increase in the accuracy achieved by inexperienced surgeons compared to freehand procedures \cite{viehofer2020augmented}.
Additional studies have explored the potential of using AR HMDs to support users during the performance of hip arthroplasties, showing comparable results to those achieved using commercial computer-assisted orthopedic systems \cite{liu2018augmented,fotouhi2018plan}.
Even more, using AR HMDs reduces the events during which the surgeons deviate their attention from the surgical scene or are exposed to fluoroscopy radiation due to unprotected parts when they turn their bodies \cite{ortega2008usefulness,fotouhi2019interactive}.

Different works have used the built-in sensors of commercially available HMDs to estimate the pose of objects of interest in medical applications.
These approaches avoid incorporating external tracking systems and additional devices into the already cluttered OR.
On the one hand, identifying markers in the visible spectrum using the built-in RGB cameras has been used to investigate the feasibility of using AR in surgical scenarios \cite{liu2018augmented,liebmann2019pedicle}.
The identification of this type of marker is not only limited to the visible spectrum and has also been investigated using multimodal images \cite{andress2018fly}.
However, the low accuracy of the registration and motion tracking may hinder their successful integration into the surgical workflow \cite{liebmann2019pedicle}.
On the other hand, the use of infrared cameras to track retro-reflective markers has been adopted by surgical grade tracking systems as they fulfill the accuracy requirements to provide surgical navigation.
While commercially available HMDs frequently incorporate infrared cameras for self-localization, hand tracking, or depth estimation, navigation accuracy analysis with these cameras requires further investigation.

One of the initial works proposing the use of the built-in cameras of AR HMDs for the detection and tracking of retro-reflective markers was proposed by Kunz et al. \cite{Kunz:2020:IR_HoloLens_Neuro}. 
This work introduced two different approaches to detecting the markers using the built-in sensors of commercially available AR HMDs (i.e., Microsoft HoloLens 1).
A first approach combined the reflectivity and depth information collected using the near field ToF sensor.
A second approach used the frontal environmental grayscale cameras as a stereo pair to estimate the position of the objects of interest.
This last approach required the addition of external infrared LEDs to illuminate the markers and increase their visibility in the grayscale images.
However, using the environmental cameras provide a higher resolution than the depth camera used as part of the ToF sensor (648x480 vs. 448x450 pixels, respectively).
Although an offline accuracy evaluation of both approaches is presented in this work, the tracking and implementation of the system are reported only for the depth-based approach, showing a tracking accuracy of 0.76 mm when the markers are placed at distances ranging between 40 to 60 cm.
In addition, only translational and not rotational experiments were reported in this work.

Additional work presented by Gsaxner et al. \cite{Gsaxner:2021:IO_HoloLend_Navigation} used the frontal environmental grayscale cameras of the Microsoft HoloLens 2 for similar purposes. 
This work introduced two tracking pipelines that use stereo vision algorithms and Kalman filters, respectively, to track retro-reflective markers in the context of surgical navigation.
The proposed framework uses the stereo vision pipeline to initialize the tracking of the markers and a recursive single-constraint-at-a-time extended Kalman filter to keep tracking of the object of interest.
Whenever the tracking using the Kalman filter is lost, the stereo vision pipeline is used to re-initialize the tracking of the objects.
This work showed that the combination of these tracking pipelines enables real-time tracking of the markers with six degrees of freedom and an accuracy of 1.70 mm and 1.11$^\circ$.
Unlike Kunz et al. \cite{Kunz:2020:IR_HoloLens_Neuro}, this work investigated the tracking errors in the position and orientation of the objects of interest.
However, this system also required the integration of additional infrared LEDs to illuminate the retro-reflective markers.
\section{Methods}\label{sec:Methods}
This work presents a framework that uses the built-in cameras integrated into commercially available AR HMDs, to enable the tracking of retro-reflective markers without adding any external components.
This section describes the necessary steps to calibrate these cameras, enable the detection of retro-reflective markers commonly attached to surgical instruments and robotic devices, identify multiple tools in the scene, and provide visual guidance to users or AR applications.
To exemplify the potential of our framework, we present a use case in which the tool tracking results serve to provide visual navigation to users during the placement of pedicle screws in orthopedic surgery. 

    \subsection{System Setup} \label{ssec:Methods_SystemSetup}
    The proposed framework was implemented using the Microsoft HoloLens 2 and a workstation equipped with an Intel Xeon(R) E5-2623 v3 CPU, an NVIDIA Quadro K4200 GPU, and 104 Gigabytes of 2133 MHz ECC Memory.
    The research mode provided for the HoloLens served to gain access to the device's built-in sensors and cameras \cite{Ungureanu:2020:HoloLens_Development}.
    To optimize the acquisition and processing of the images collected using the HMD, we connected this device to the workstation using a USB-C to ethernet adapter and a cable that provided a 1000 Mbps transmit rate.
    A multi-threaded sensor data transfer system, implemented in Python via TCP socket connection, enabled the acquisition of the sensors' data at a high frame rate and with low latency.
    The data transfer system was tested using two different approaches.
    A first approach, using DirectX, allowed to transfer and acquire the video from the cameras at an approximate framerate of 31 fps.
    A second approach, implemented using Unity 3D, reported an average speed of 21 fps.
    In addition, a user interface was designed to collect, process, and analyze the data collected from the HMD.
    This user interface also provided visual guidance and assisted users while performing tasks requiring the precise alignment of virtual and real objects in surgical scenarios.

    \subsection{Camera Calibration and Registration} \label{ssec:Methods_CameraCalibration}
    \textbf{\textit{Camera Calibration.}} To investigate the feasibility of utilizing and combining the multiple cameras integrated into the HoloLens 2 for the tracking of retro-reflective markers, the left front (LF), right front (RF), main (RGB), and Articulated HAnd Tracking (AHAT) cameras need to be calibrated.
    This calibration process enables extracting the spatial relationship between the cameras and their intrinsic parameters and representing them using a unified coordinate system.
    
    We used a checkerboard with a $9\times12$ grid containing individual squares of $2cm$ per side to calibrate the different sensors.
    The checkerboard was attached to a robotic arm, and the HoloLens was rigidly fixed on a head model to ensure steadiness during image capturing.
    A set of 120 pictures, acquired by placing the checkerboard at different positions and orientations using the robotic arm, were collected for each one of the cameras.
    The images for the RGB camera were captured using the HoloLens device portal with a resolution of $3904\times2196\ pixels$, while the other cameras' images were retrieved via the HoloLens research mode.
    For the AHAT camera, the calibration and registration procedure was completed using the reflectivity images.
    
    After data collection, the images were randomly subdivided into two subsets. 
    The first subset contained 90 pictures used for camera calibration, and the remaining 30 images served for testing.
    The Root Mean Square Error (RMSE) of the detected and reprojected corner points was used to evaluate the calibration and test image sets.
    The images selection process was repeated twenty times to ensure the repeatability of the results.
    The reprojection errors for the calibration and test sets resulted in $0.165\pm 0.032\ pixels$ and $0.152\pm 0.052\ pixels$ for the AHAT, $0.069\pm 0.001\ pixels$ and $0.070\pm 0.002\ pixels$ for the LF, $0.080\pm 0.002\ pixels$ and $0.080\pm 0.002\ pixels$ for the RF, and $0.305\pm 0.020\ pixels$ and $0.308\pm 0.060\ pixels$ for the RGB, respectively.
    
    \noindent \textbf{\newline \textit{Camera Registration.}} For the registration of the different cameras, a new set of 100 images were collected from the different sensors while placing the checkerboard at different poses.
    The checkerboard was kept steady before moving it to a different position to ensure synchronization between the multiple sensors.
    As an initial step, the RF camera was registered to the LF camera using the \textit{stereo camera calibration} toolbox from Matlab.
    The following step registered the LF and RGB cameras to the AHAT.
    For the registration of these cameras, the three-dimensional positions of the checkerboard's corner points were calculated using the respective intrinsic camera parameters.
    The extrinsic parameters between two cameras were later calculated by solving a least-squares optimization problem:
    \begin{equation}
        (R_c^A,t_c^A)=\mathop{argmin}\limits_{R\in SO(3), t\in R^3}\sum_{i=1}^{n}||(Rp_i+t)-q_i||^2
        \label{equation:LeastSquareOptimization}
    \end{equation}
    \noindent where $p_i$ represents the three-dimensional corner points in the RGB and LF cameras, and $q_i$ denotes the corner points in AHAT space. 
    
    The registration error was evaluated using the RMSE between the detected corner points of one camera and the reprojected points of the other camera. 
    The registration between the RF-LF cameras produced a $0.190\ pixels$ reprojection error.
    In comparison, the registration between the LF-AHAT and RGB-AHAT cameras led to reprojection errors of  $0.901$ and $0.758\ pixels$, respectively.
    
    In addition, the field-of-view (FoV) of every camera was estimated quantitatively in horizontal and vertical directions using the equations:
    \begin{equation}
        FoV_x = \max_{0<i<j\le n} arccos\left(\frac{(x_i,0,1)\cdot(x_j,0,1)}{||(x_i,0,1)||\times ||(x_j,0,1)||}\right)
        \label{equation:FoVCalculationx}
    \end{equation}
    \begin{equation}
        FoV_y = \max_{0<i<j\le n} arccos\left(\frac{(0,y_i,1)\cdot(0,y_j,1)}{||(0,y_i,1)||\times ||(0,y_j,1)||}\right)
        \label{equation:FoVCalculationy}
    \end{equation}
    
\begin{figure}[t]
        \centering
        \includegraphics[page=3,width=0.93\columnwidth]{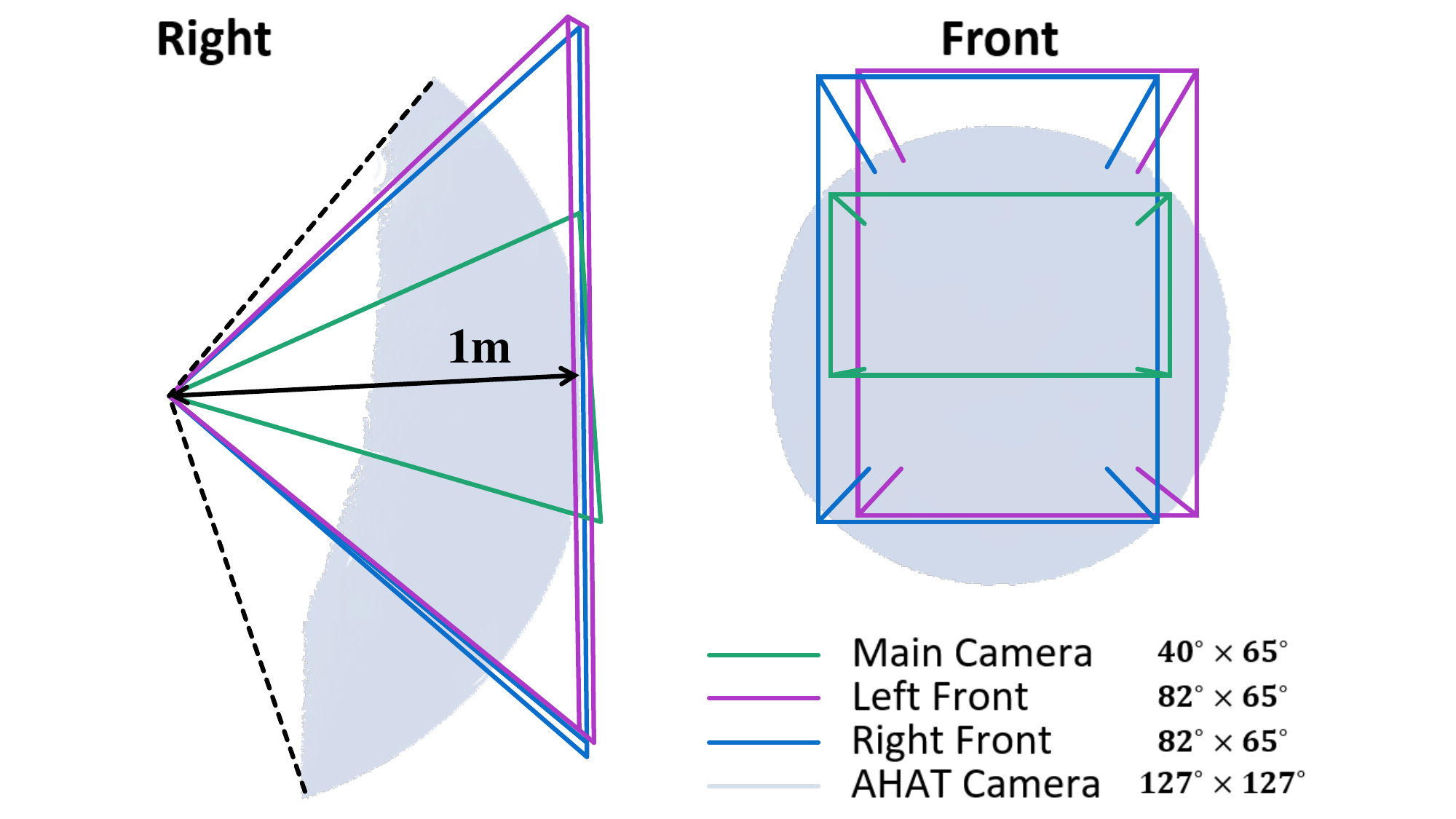}
        \caption{The spatial relationship between the built-in cameras of the HoloLens 2. The camera spaces of the RGB, LF, and RF cameras are displayed using frustums. The camera space of the AHAT camera is presented using a one-meter radius sphere. The FoV of all the cameras is depicted as $height \times width$.}
        \label{fig:3Dspace}
    \end{figure}

    Once the registration between the cameras is complete, the different sensors can be presented in a common coordinate space using their corresponding camera parameters (see Figure~\ref{fig:3Dspace}).
    This figure depicts the FoV of the visible light cameras from the optical center to a unit plane using frustums.
    As the AHAT sensor data is only effective within a 1-meter distance, the FoV of this sensor is depicted by the furthest detection surface.
    
    During the calibration process presented in this work, a $12.76^\circ$ angle down was observed for the AHAT compared to the other cameras.
    The FoV of the AHAT camera resulted in $127^\circ \times 127^\circ$, which is about six times larger than the RGB camera ($40^\circ \times 65^\circ$) and three times larger than the LF and RF cameras ($82^\circ \times 65^\circ$), making it extremely useful for the target detection within a meter distance, where most of the actions in the personal space take place. 
    In addition, the relatively large FoV of the AHAT camera, combined with the sensor information of environmental infrared reflectivity and depth, makes this camera suitable for detecting retro-reflective markers used in surgical procedures.
    
    \subsection{Tool Tracking} \label{ssec:Methods_ToolTracking}
\begin{figure}[!ht]
        \centering
        \includegraphics[page=1,width=0.9\columnwidth]{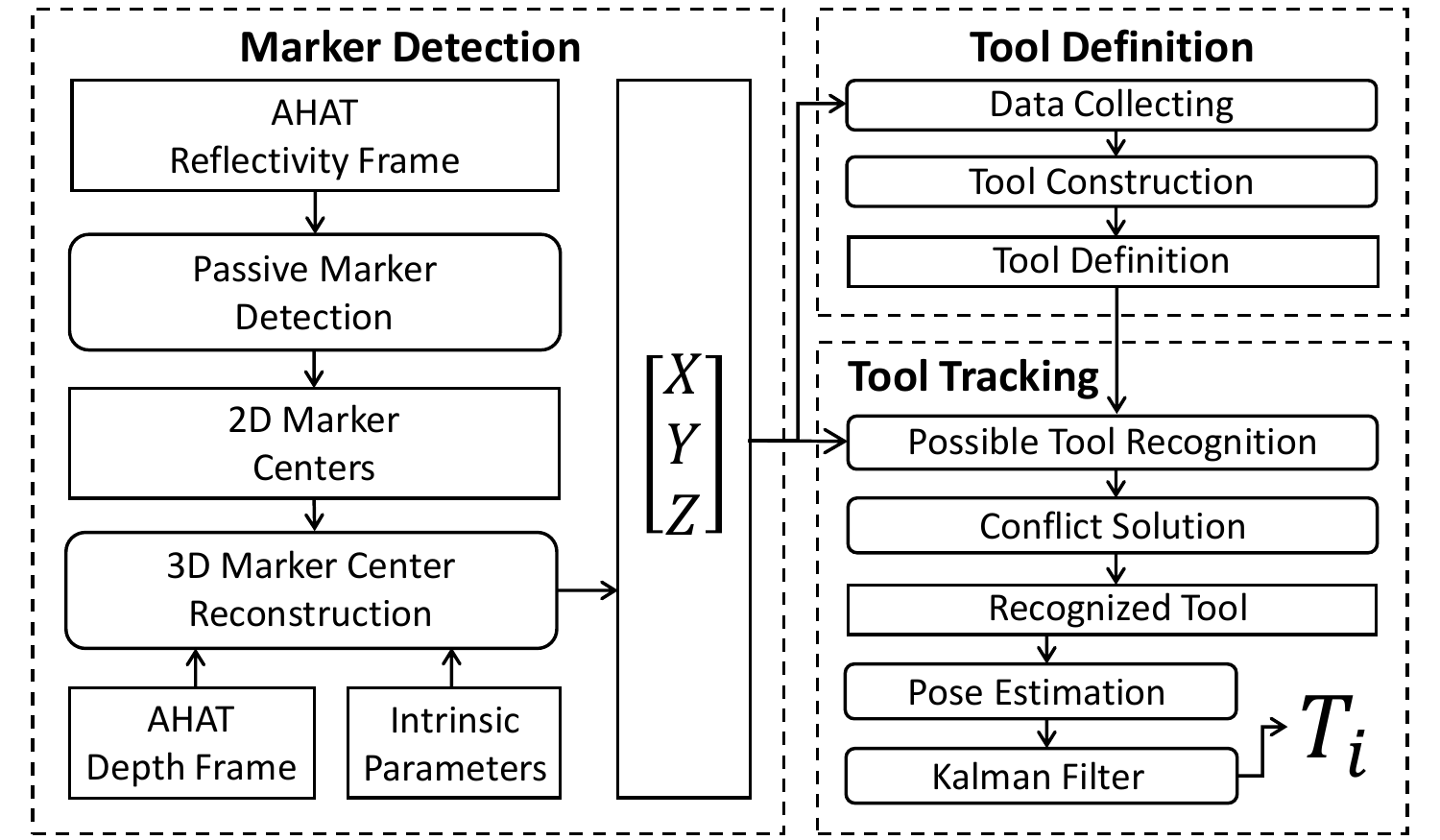}
        \caption{The proposed framework comprises three main stages. i) The three-dimensional position of all the retro-reflective markers observed using the AHAT camera is extracted from the scene. ii) The detected markers are grouped in subsets to identify particular arrangements corresponding to specific tools. iii) The pose of the identified tools is extracted from subsequent image frames.}
        \label{fig:ToolTrackingFlow}
    \end{figure}

    To explore the capabilities of the AHAT camera to track retro-reflective markers used in surgical scenarios, we propose a framework for the definition, recognition, and localization of tools that use this specific type of marker.
    As depicted in Figure~\ref{fig:ToolTrackingFlow}, such a framework includes three main stages: three-dimensional marker's center detection, tool definition, and multi-tool recognition and localization.

        \subsubsection{Marker Detection} \label{sssec:Methods_MarkerDetection}
        As an initial step, the three-dimensional positions of the individual retro-reflective markers are obtained using the intrinsic camera parameters and the reflectivity and depth images of the AHAT camera.
        This type of retro-reflective marker depicts extremely high-intensity values when observed in the reflectivity images of the AHAT camera.
        Therefore, an intensity-based threshold criterion was used to separate these objects of interest from the rest of the scene.
        The AHAT intensity images are represented using a 16-bit unsigned integer format.
        As depicted in Figure~\ref{fig:Pipeline}, most of the environment presents a reflectivity intensity below 300.
        At the same time, the retro-reflective markers report values larger than 500, with a peak value over 2000 at the center of the marker.
        
        In addition, a connected component detection algorithm was used to separate individual markers from each other.
        Considering the environmental noise observed in the reflectivity images, including large connected high-reflection areas caused by flat reflective objects like glass or monitors or smaller areas resulting from random noise or surrounding objects, an additional threshold on the connected component area was applied to extract the individual retro-reflective markers. 
        To estimate the area in pixels $A_{px}$ that a retro-reflective marker with radius $r$ at a distance $d$ would occupy in the image of the AHAT camera, we used the following equation:
        \begin{equation}
            A_{px}\approx \frac{\pi \cdot r^2}{s_x/f_x \cdot s_y/f_y \cdot d^2}
            \label{equation:pixelsize}
        \end{equation}
        \noindent where $s_x$, $s_y$, $f_x$, and $f_y$ are the pixel size in \textit{mm/px} and the focal distance in \textit{mm} of the AHAT intrinsic parameters.
        
        After applying the intensity-based and connected component detection algorithms, the three-dimensional position of every retro-reflective marker is calculated using the AHAT camera's depth information $d_i$, its intrinsic parameters, and the marker's diameter.
        The central pixel of every marker is first back-projected into a unit plane $(x_i,y_i)$ using the intrinsic parameters. 
        The depth information from the AHAT camera represents the distance between a certain point in the three-dimensional space and the camera's optical center\cite{Labini:2019:Depth_awareness}.
        Hence, the position $(X_{i},Y_{i},Z_{i})$ of this point can be expressed as follows:
        \begin{equation}
            \left(X_{i},Y_{i},Z_{i}\right)=\frac{d_{i}}{||(x_{i},y_{i},1)||_2}\left(x_{i},y_{i},1\right)
            \label{equation:3Dpositionreconstruction}
        \end{equation}
        While the calculated position corresponds to the actual center of the target when using flat markers, the radius of the target must be considered when using spherical markers.
        In this particular case, the estimated depth information provided by the AHAT camera corresponds to the central point on the surface of the marker and not to the center of the sphere.
        Therefore, when using spherical markers, the central position of the sphere can be computed as:
        \begin{equation}
            \left(X_{i},Y_{i},Z_{i}\right)_{sphere}=\frac{d_{i}+r}{d_{i}}\left(X_{i},Y_{i},Z_{i}\right)
            \label{equation:3DpositionreconstructionSphere}
        \end{equation}
        \noindent where $r$ represents the radius of the spherical marker.

        \subsubsection{Tool Definition} \label{sssec:Methods_ToolDefinition}
        A common approach to tracking objects using retro-reflective markers involves attaching a set of these components to a rigid frame with a unique spatial distribution.
        This spatial information contributes to estimating the pose of a particular object and allows distinguishing between different instances that can be observed simultaneously in the scene.
        The procedure to define the properties of an object using our framework, from now on referred to as a \textit{tool}, requires a series of steps that are described next.

        First, a single image frame from the AHAT camera is used to extract the three-dimensional distribution of the retro-reflective markers that belong to a tool.
        The spatial distribution of such markers provides the geometrical properties of the tool and enables the generation of an initial configuration of it.
        This step allows identifying potential misdetection in subsequent frames by comparing the distance observed between the detected markers.
        
        The following step uses an optimization method to calculate the shape of the rigid tool using the $N$ three-dimensional positions of the markers extracted from every frame.
        In this step, we use $T_i:\{M_{i,j}(x,y,z)\}$ to represent the set of markers $T$, in the $i^{th}$ frame of the total collection, that defines the three-dimensional position of the individual markers $M$ with index $j$ observed in the tool definition.

        The RMSE between one tool in two frames, $T_p$ and $T_q$, is used to evaluate the difference as follows:
        \begin{equation}
            \Delta(T_p,T_q)=\sqrt{\frac{1}{N}\sum_{0\le j<N}||M_{p,j}-M_{q,j}||_2^2}
        \end{equation}
        where $||\cdot||_2$ is the Euclidean distance of the marker's position with index $j$ observed in the frames $p$ and $q$.

        Furthermore, to estimate the optimal position of the set of markers that define a tool, the mean difference between the optimized $T_{opt}$ and estimated $T_{i}$ positions in every frame are used as the minimization target as follows:
        \begin{equation}
            T_{opt}=\mathop{argmin}\limits_{T:\{M_{j}\}}\frac{1}{I}\sum_{0\le i<I}\Delta(T,T_i)
            \label{equation:ToolDefOptimization}
        \end{equation}
        \noindent where I represents the total number of collected frames for the tool definition procedure.

        Lastly, the mean translation of the set of markers is removed to generate the definition $T_{def}:\{M_{def,j}\}$ using:
        \begin{equation}
            M_{def,j}=M_{opt,j}-\frac{1}{N}\sum_{0\le k<N}M_{opt,k}
            \quad  ;\   0\le j < N
            \label{equation:ToolDefFinal}
        \end{equation}
        \noindent Here, $T_{def}$ represents the final definition of the tool with origin at the geometric center of all the markers in the tool.
    
        \begin{figure}[!t]
            \centering
                \includegraphics[page=1,width=\columnwidth]{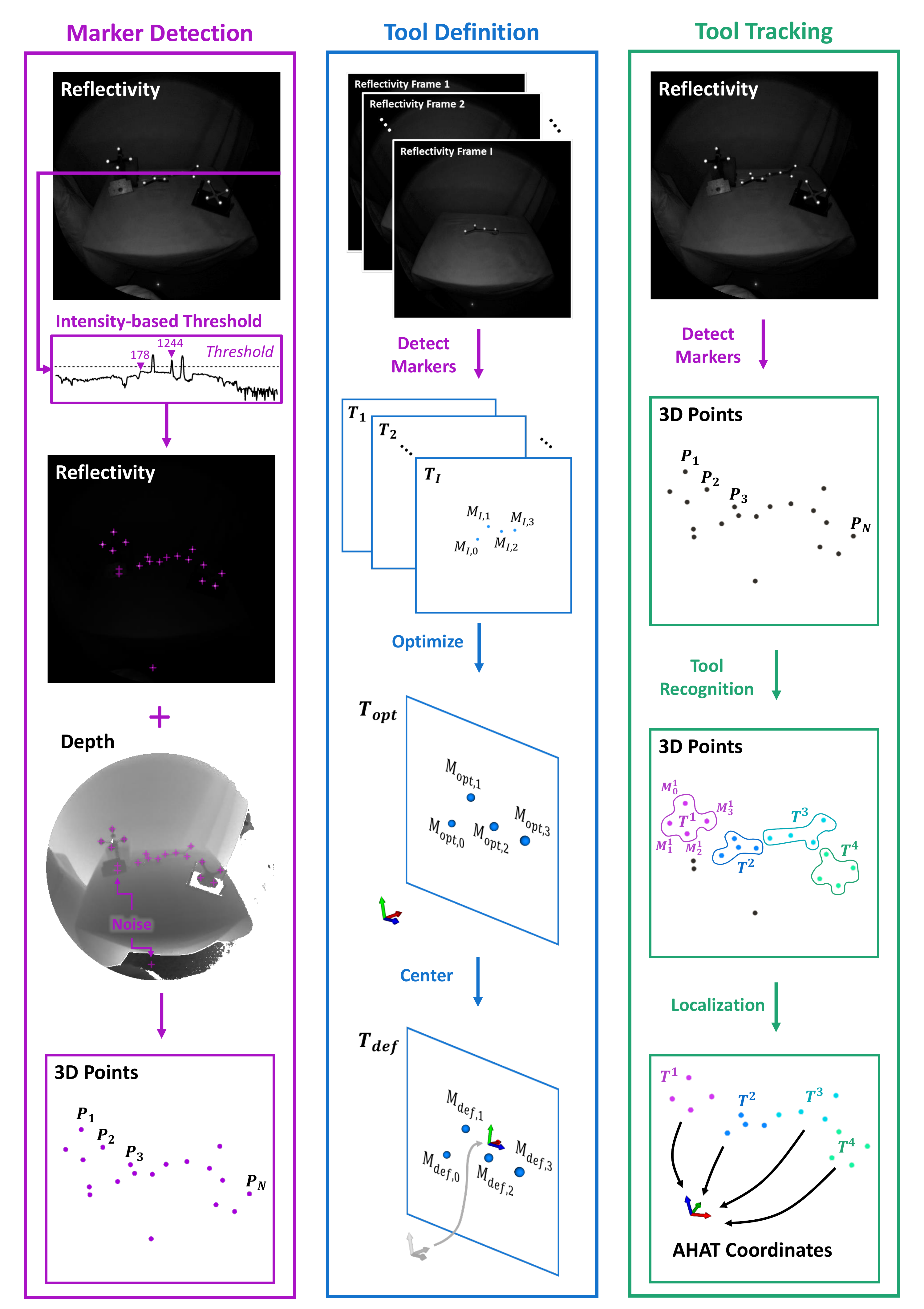}
            \caption{The proposed framework combines the reflectivity and depth images of the AHAT camera to detect and extract the three-dimensional position of passive retro-reflective markers. The marker detection stage facilitates identifying and defining individual tools composed of multiple spherical markers. This approach allows localizing and recognizing multiple tools in the scene without modifying the AR-HMD.}
            \label{fig:Pipeline}
        \end{figure}
        
        This procedure is depicted in Figure~\ref{fig:Pipeline}.

        \subsubsection{Tool Recognition and Localization}
        To identify multiple tools, each with shape $T^l : \{M_{j}^{l}\}$, from a single AHAT frame where multiple markers $\{P_n\}$ are detected, we estimate the Euclidean distance between the markers and tools using the following equations:
        \begin{equation}
            \begin{matrix}
                L_P(n,n+\alpha)=||P_n-P_{n+\alpha}||_2\quad \\ \text{for: } 0 \le n<n+\alpha<N
            \end{matrix}
        \end{equation}
        \begin{equation}
            \begin{matrix}
                L_{T^l}(n,n+\alpha)=||M_{n}^{l}-M_{n+\alpha}^{l}||_2\quad \\ \text{for: } 0\le l < L\  ;\  0 \le n<n+\alpha<N_l
            \end{matrix}
        \end{equation}
        \noindent where $n$ indicates the $n^{th}$ marker in the tool or a scene, and $l$ represents the $l^{th}$ tool to be detected in the environment.

        All subsets $\{P_n\}_{T^l,k}$ of detected markers $P$ whose shape fits the definition of the tool  $T^l$ are found using a depth-first graph searching algorithm; $k$ refers to the $k^{th}$ possible subset of markers that fits the target tool.
        To quantitatively depict the similarity between the possible solution and target tool definition, we used a loss function that considers the corresponding lengths of the tool:
        \begin{equation}
        \begin{split}
            \mathcal{L}(T_{def}^l,\{P_n\}_{T^l,k})=\frac{1}{N_l(N_l-1)}\sum_{0\le i<j<N_l} \beta_{T^l,k}(i,j)
            \\
            \beta_{T^l,k}(i,j) = || L_{T_{def}^l}(i,j)-L_{\{P_n\}_{T^l,k}}(i,j) ||
        \end{split}
        \end{equation} 
        \noindent where $N_l$ refers to the number of markers in the definition of the $l^{th}$ tool.

        During the searching process, we define two thresholds to exclude incorrect matches.
        A first threshold $t_{side}$ identifies mismatches between the corresponding sides of the tool.
        A second threshold $t_{shape}$ discerns between the obtained solution and the tool definition.
        These thresholds are applied as follows:
        \begin{equation}
            t_{side} > \beta_{T^l,k}(i,j)\quad  ;\  0\le i<j<N_l
        \end{equation}
        \begin{equation}
            t_{shape} > \mathcal{L}(T_{def}^l,\{P_n\}_{T^l,k})
        \end{equation}
        To select a proper value for these thresholds, the uncertainty of the three-dimensional position of the detected markers $P_n$ is considered.
        As shown in Figure~\ref{fig:Gauss_Test_Scene_C}, the standard error for depth detection, $\sigma_{p}$, changes as a function of the depth at which the markers are observed.
        This value serves to calculate the maximum standard error for a side using $\sigma_{side} = \sqrt2\sigma_p$.
        When a $95\%$ probability for detection is desired, the thresholds for a single side can then be calculated as $t_{side}=2\sigma_{side}$.
        Lastly, the error threshold for the whole tool is computed using  $t_{shape}=t_{side}/\sqrt{N_{l}(N_{l}-1)}$.

        After the possible solutions for every tool have been extracted, the following step deals with any redundant information found in the possible solutions.
        This redundant information includes using the same retro-reflective marker more than once or using the same set of markers to track the same tool.
        When multiple solutions use the same set of markers, ordered using different index values, to estimate the pose of the same tool, the solution with the lowest error $\mathcal{L}$ is preserved, and the remaining solutions are removed.
        After removing these redundant solutions, the remaining subsets are sorted according to their error $\mathcal{L}$ using their respective tool definition $T^l$.
        In this case, the solution with the least error is considered the reference.
        Any remaining solution that conflicts with the reference or aims at detecting the same tool is removed.
        This procedure is repeated until all the defined tools are found or no other solution exists.
        A final step uses an SVD algorithm to calculate the transformation matrices that map the tool coordinates into the AHAT camera space.

        \subsubsection{Precision Enhancement}

        \begin{figure*}[htpb]
            \centering
                \subfloat[\label{fig:Gauss_Test_Scene_A} Experimental Setup]{\includegraphics[page=1,width=0.65\columnwidth]{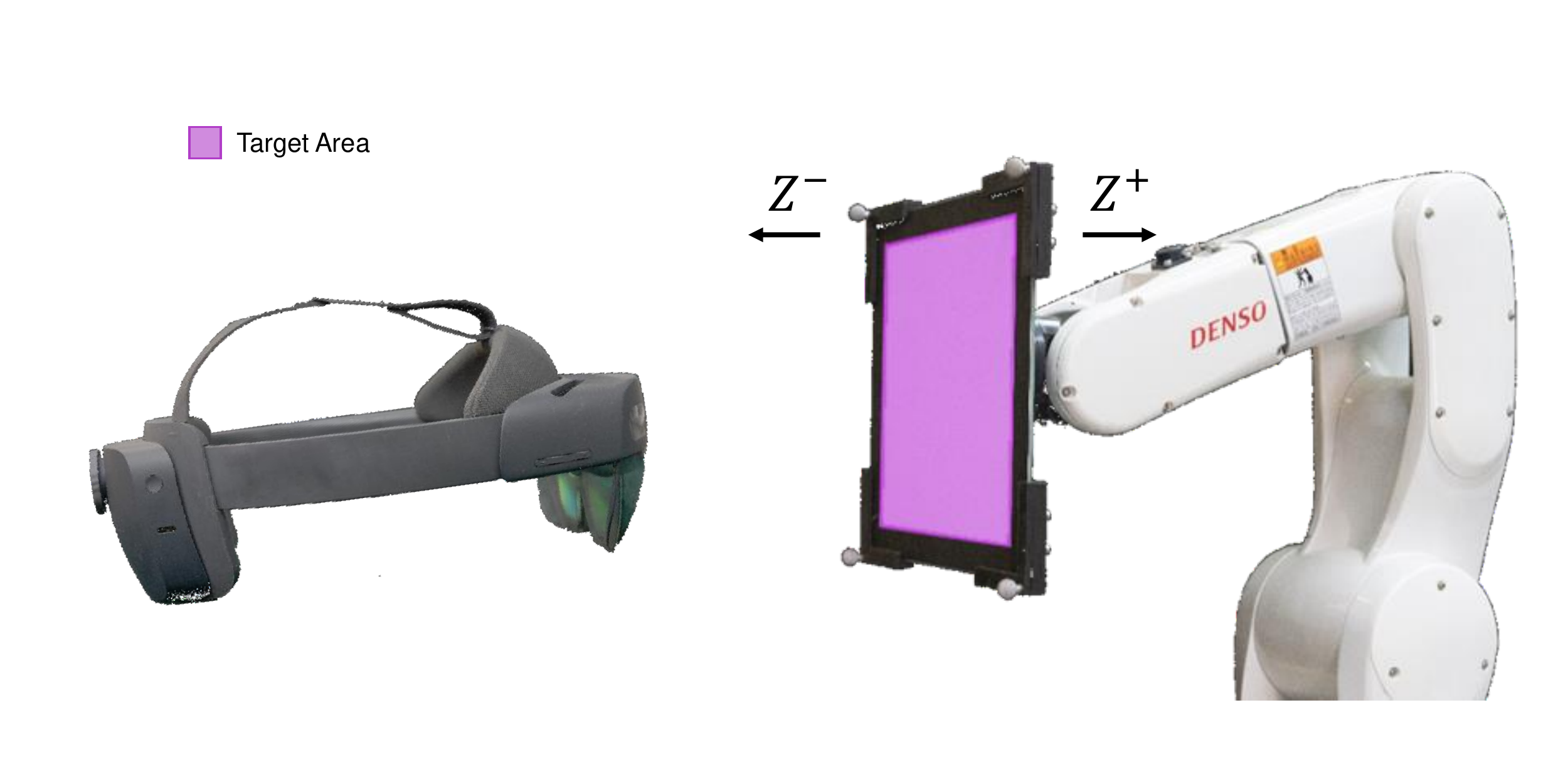}}\hfill
                \subfloat[\label{fig:Gauss_Test_Scene_B} Point-to-plane Distance (mm)]{\includegraphics[page=2,width=0.65\columnwidth]{Figures/GaussKalman_v2.pdf}}\hfill
                \subfloat[\label{fig:Gauss_Test_Scene_C} Depth (mm)]{\includegraphics[page=5,width=0.65\columnwidth]{Figures/GaussKalman_v2.pdf}}
            \caption{\protect\subref{fig:Gauss_Test_Scene_A} Experimental setup used to estimate the noise distribution provided by the built-in depth camera on the HMD. \protect\subref{fig:Gauss_Test_Scene_B} The point-to-plane distance error distribution at approximately 200mm, 500mm, and 700mm depths. \protect\subref{fig:Gauss_Test_Scene_C} The standard error of depth detection for every pixel versus depth. The vertical lines depict the mean depth at which the target was detected.}
            \label{fig:Gauss_Test_Scene}
        \end{figure*}

        In addition, we proposed using Kalman filters to enhance the precision that can be achieved with the proposed framework.
        On the one hand, using these filters could contribute to reducing the tremor observed by the cut-off error from the AHAT depth data.
        This source of error is observed because the AHAT provides an integer value corresponding to the depth detected in millimeters, limiting the maximum depth resolution to 1 mm and decreasing the tracking stability of the passive markers and tools.
        On the other hand, the implementation of Kalman filters could compensate for the detection error observed when using the AHAT sensor.
        However, to justify the use of Kalman filters, it is required that the signals involved follow a gauss distribution.
        This type of distribution allows the filters to behave as a best linear unbiased estimator.
        Therefore, we conducted an experiment to estimate if the AHAT's detection error can be modeled using a gaussian distribution.

        Our experimental setup used a glass board with an anodized aluminum surface and retro-reflective markers located at the corners.
        A robotic arm was used to move the glass board along the \textit{z-axis} of the AHAT camera frame while keeping the board steady during data collection (Fig.~\ref{fig:Gauss_Test_Scene_A}).
        The error distribution of the depth data was tested using 48 different depths, assigned randomly and ranging from 156 to 971 mm.
        These lower and upper boundaries correspond to the limits where the board occupies the whole image until the depth data is invalid.
        For every depth distance, 300 continuous AHAT frames were collected.
        The reflectivity information of the first frame was used to extract the target detection area.
        Every depth image contained in the 300 frames was used to create individual point clouds merged to produce a final set.
        A plane fitting algorithm used the merged point cloud to estimate the board's position.
        
        Finally, the fitting error was estimated using a point-to-plane function distance.
        As depicted in Figure~\ref{fig:Gauss_Test_Scene_B}, the error presents a strong gauss-like distribution at different depths.
        An Anderson-Darling test was used to test normal distribution, where the \textit{p-value} was found to be smaller than 0.0005 for all the 48 distances. 
        In addition, the standard deviation increased as a function of the depth.
        Further details regarding the error observed in the data at the different depths are presented in Figure~\ref{fig:Gauss_Test_Scene_C}.
        This relationship can be modeled using a quadratic polynomial with a coefficient of determination equal to 0.9807.
        
        These results support the use of a Kalman filter for the proposed framework.
        Thus, we used several independent filters for the individual markers for every tracked tool based on their depth value.
        After the filtered depth value is calculated, the three-dimensional position and transform matrix from the tool to AHAT space are adjusted.
        In addition, when the tool's tracking is lost, every Kalman filter for the specific tool is reinitialized.
\section{Experiments and Results}\label{sec:result}

\subsection{Localization Error}
To assess the tracking accuracy of our algorithm, we conducted a set of experiments in which we compared different tracking technologies in a controlled environment.
Among these tracking technologies, we used optical cameras to detect ArUco and ChArUco markers in the visible spectrum.
Although existing works have used the environmental cameras of the HoloLens for the tracking of these markers \cite{gibby2019head}, we used an MSIP-RM-PGR-CMU313Y3 camera\footnote{FLIR Integrated Imaging Solutions, Inc.} coupled with an M1614-MP2 lens\footnote{CBC AMERICA LLC.}.
This camera provided a resolution of $1280\times 1024$ pixels and enabled the acquisition of images with better image quality than those acquired using the built-in cameras of the HoloLens.
In addition, we compared the performance of our algorithm against an NDI Polaris Spectra\footnote{Northern Digital Incorporated}.
This system is commonly used in surgical procedures to track passive retro-reflective markers in the infrared spectrum.

\subsubsection*{Stability, Repeatability, and Tracking Accuracy}
To compare the stability, repeatability, and accuracy of the optical markers' pose estimation using different tracking technologies, we conducted an experiment in which we translated and rotated the target markers using six different configurations.
These experiments targeted multiple distances in the personal space at reaching distances.
To precisely control the movements of the markers for the different sensors, we used a \textit{3-axis} linear translation stage and a rotation platform with an accuracy of $0.01mm$ and $0.1^{\circ}$, respectively.
The markers were left static for the first experiment, and 100 poses were collected.
After data collection, the markers were moved along the \textit{x-axis} by $1 mm$, and another set of 100 poses was recorded.
The following step brought back the markers to their original position and a total number of 10,000 randomly selected distance values were calculated using these two data sets.
We performed this procedure 20 times using different initial marker poses and environmental light conditions to ensure repeatability.
A second experiment followed this procedure but used a translation of $20 mm$ along the \textit{x-axis}.
The markers were moved along the \textit{z-axis} using the same translations for the third and fourth experiments.
Lastly, the markers were rotated 10 and 50$^{\circ}$ around the \textit{world-up} vector for the fifth and sixth experiments.

\begin{figure}[!b]
    \centering
        \includegraphics[page=1,width=0.49\columnwidth]{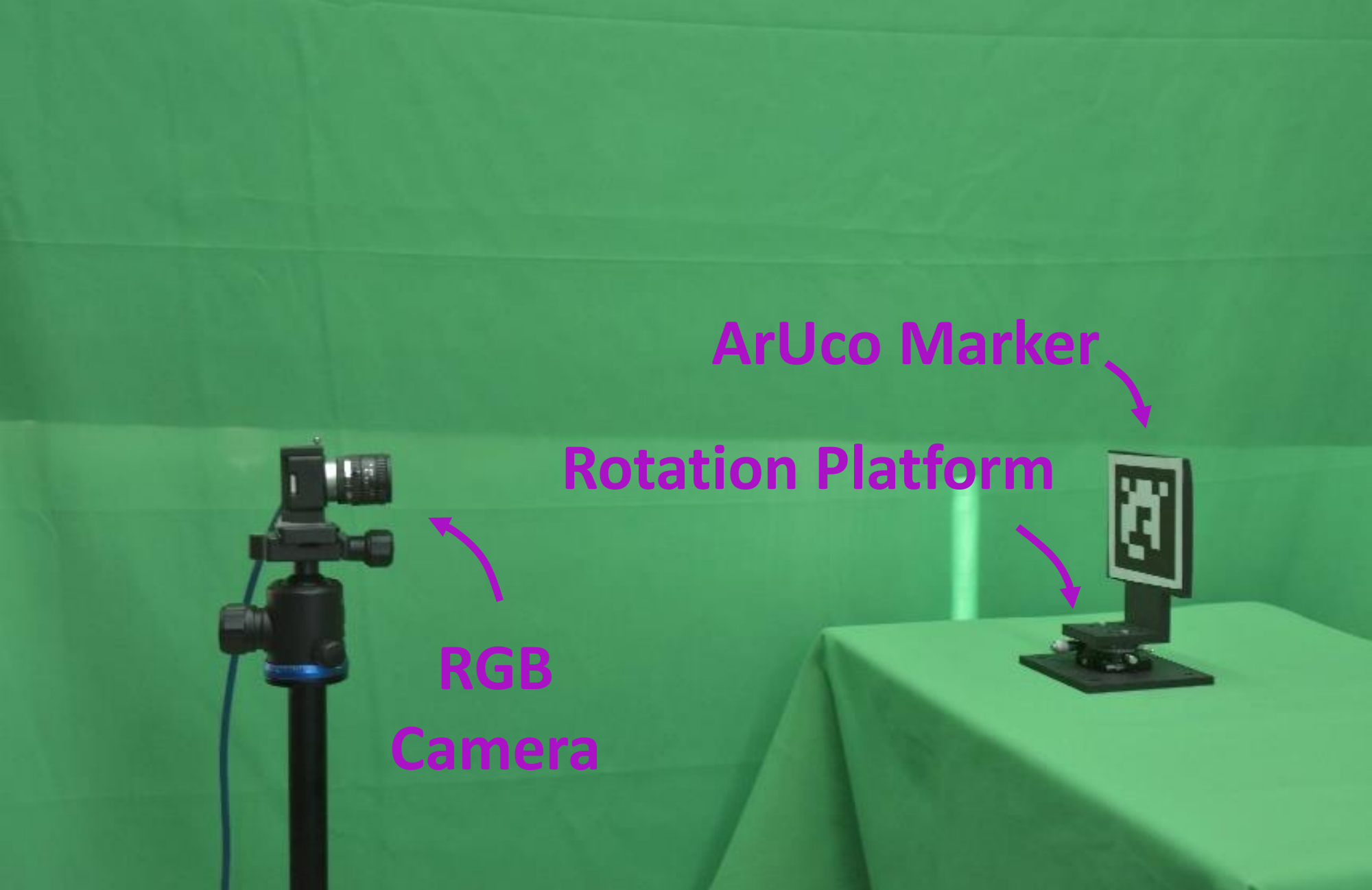}\hfill
        \includegraphics[page=2,width=0.49\columnwidth]{Figures/Markers.pdf}
    \vspace*{-3mm}
    \caption{Experimental setup for tracking accuracy using ArUco and ChArUco (\textit{left}) and retro-reflective (\textit{right}) markers.}
    \label{fig:markers}
\end{figure}

\begin{figure*}[t!]
	\begin{center}
		\begin{tabular}{ccccc}
			\raisebox{-\totalheight}{\includegraphics[page=1,width=0.3\textwidth]{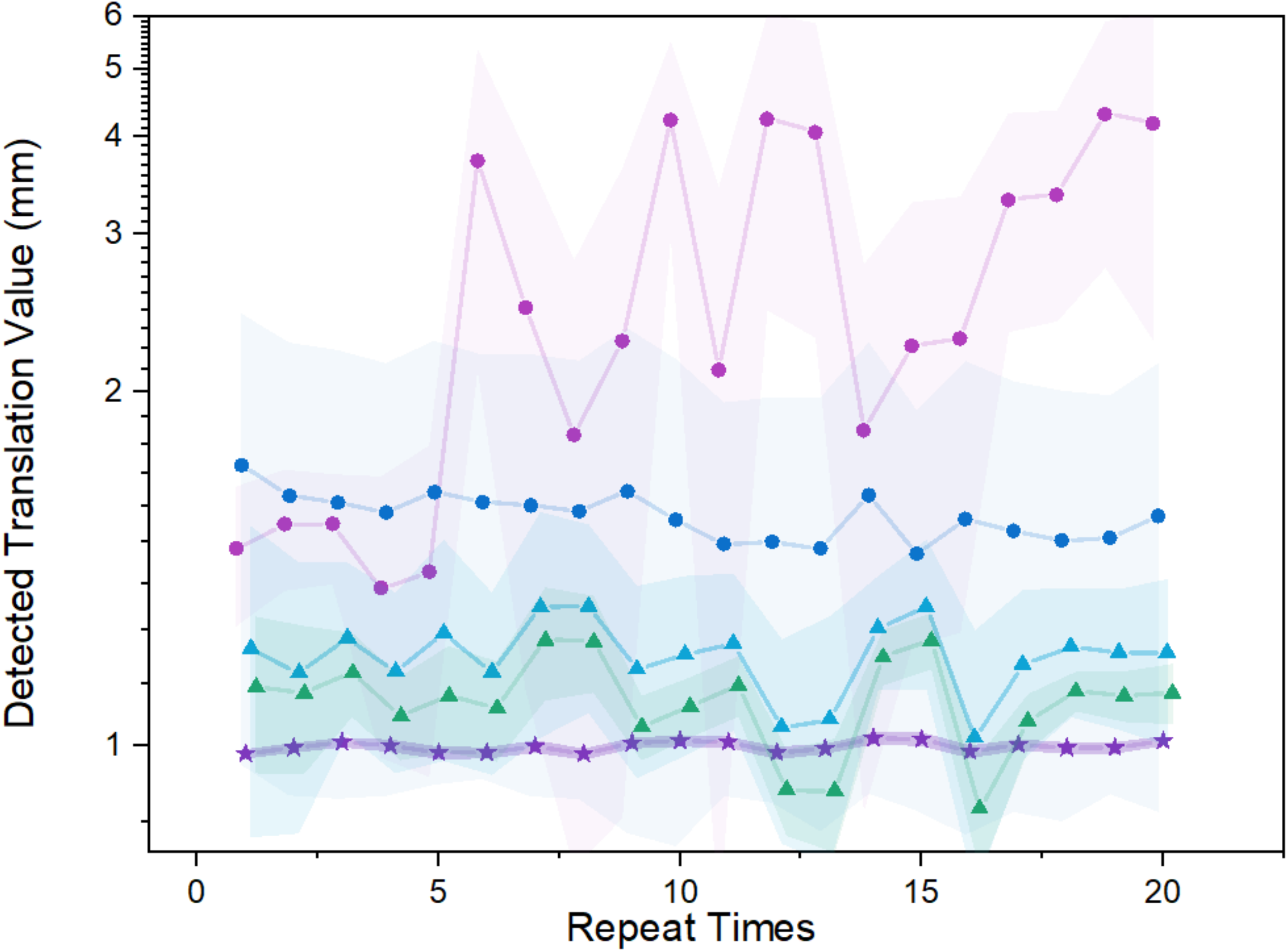}}
			& &
			\raisebox{-\totalheight}{\includegraphics[page=3,width=0.3\textwidth]{Figures/Precision_v2.pdf}}
			& &
			\raisebox{-\totalheight}{\includegraphics[page=5,width=0.3\textwidth]{Figures/Precision_v2.pdf}}
			\\
			\scriptsize{(a) 1 mm translation \textit{x-axis}} & & \scriptsize{(b) 1 mm translation \textit{z-axis}} & & \scriptsize{(c) $10^\circ$ rotation}
			\\ 
			\cmidrule(l){1-1}\cmidrule(l){3-3}\cmidrule(l){5-5}
			\raisebox{-\totalheight}{\includegraphics[page=2,width=0.3\textwidth]{Figures/Precision_v2.pdf}}
			& &
			\raisebox{-\totalheight}{\includegraphics[page=4,width=0.3\textwidth]{Figures/Precision_v2.pdf}}
			& &
			\raisebox{-\totalheight}{\includegraphics[page=6,width=0.3\textwidth]{Figures/Precision_v2.pdf}}
			\\
			\scriptsize{(d) 20 mm translation \textit{x-axis}} & & \scriptsize{(e) 20 mm translation \textit{z-axis}} & & \scriptsize{(f) $50^\circ$ rotation}
			\\ 
			\cmidrule(l){1-1}\cmidrule(l){3-3}\cmidrule(l){5-5}
			\multicolumn{5}{c}{\raisebox{-\totalheight}{\includegraphics[page=1,width=0.75\textwidth]{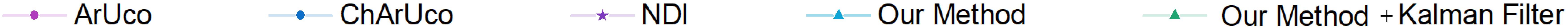}}}
		\end{tabular}
		\caption{Experiment results of tool tracking accuracy assessment of our infrared depth camera and Kalman filter-based tracking method, compared with commercial IR and visible light marker tracking methods. (a),(d) Distribution of detected moving distances when moving optical platform 1 and 20 mm along \textit{x-axis} over 20 repetitive tests. (b),(e) Distribution of detected moving distances when moving the optical platform 1 and 20 mm along the \textit{z-axis}. (c),(f) Distribution of detected rotation angle when rotating the optical platform 10 and 50 degrees.}
    \label{fig:precision}
	\end{center}
\end{figure*}

To ensure that the tracking systems could provide accurate values, we placed the markers at different distances from the sensors depending on the tracking technology used.
For our proposed method, a tool composed of four retro-reflective spheres with a diameter of $11.5 mm$  was placed at an approximate distance of $600 mm$ from the AHAT camera.
The same tool was placed at approximately $2000 mm$ for the NDI tracking system.
In addition, for the detection of markers in the visible spectrum, we used a $6\times6$ ArUco marker and a $3\times3$ ChArUco marker, each with a side length of $80 mm$.
These optical markers were placed at $800 mm$ from the camera.
The experimental setup used for tracking these targets is shown in Figure~\ref{fig:markers}, and the results of this experiment are presented in Figure~\ref{fig:precision}.

\begin{figure}[t!]
	\begin{center}
		\begin{tabular}{ccc}
			\raisebox{-\totalheight}{\includegraphics[page=7,width=0.2\textwidth]{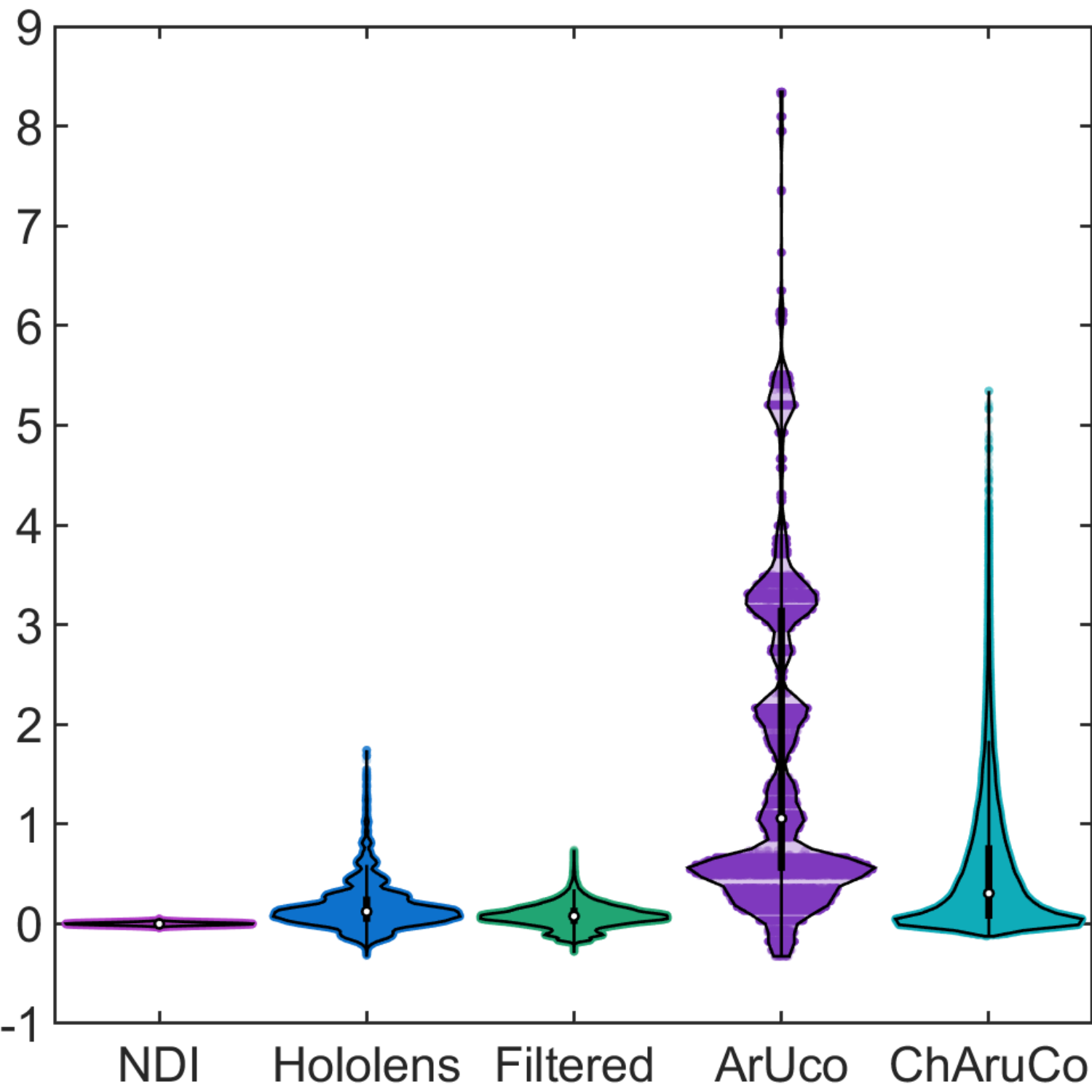}}
			& &
			\raisebox{-\totalheight}{\includegraphics[page=8,width=0.2\textwidth]{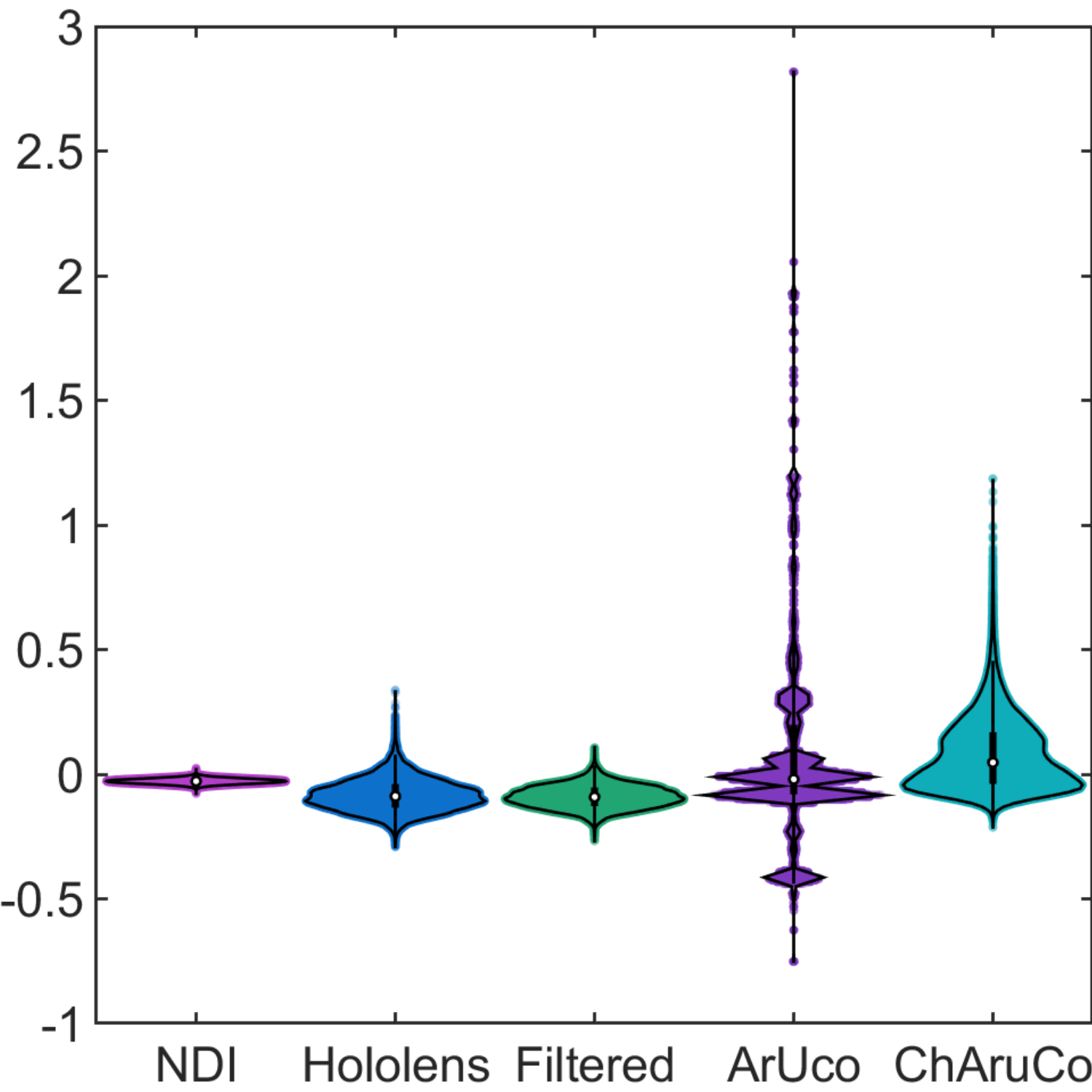}}
			\\
			\scriptsize{(a) 1 mm translation \textit{x-axis}} & & \scriptsize{(b) 20 mm translation \textit{x-axis}} \\ 
			\cmidrule(l){1-1}\cmidrule(l){3-3}
			\raisebox{-\totalheight}{\includegraphics[page=9,width=0.2\textwidth]{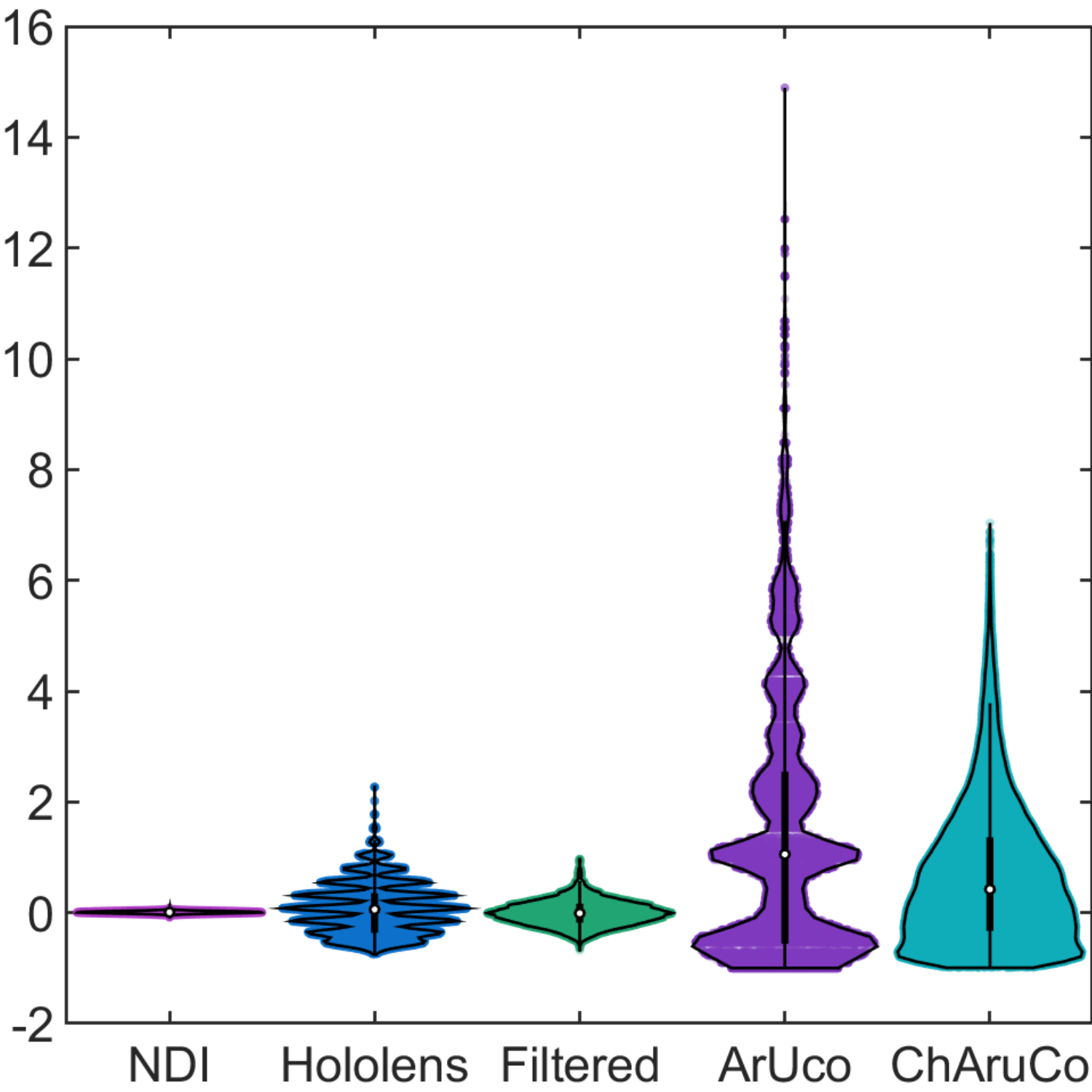}}
			& &
			\raisebox{-\totalheight}{\includegraphics[page=10,width=0.2\textwidth]{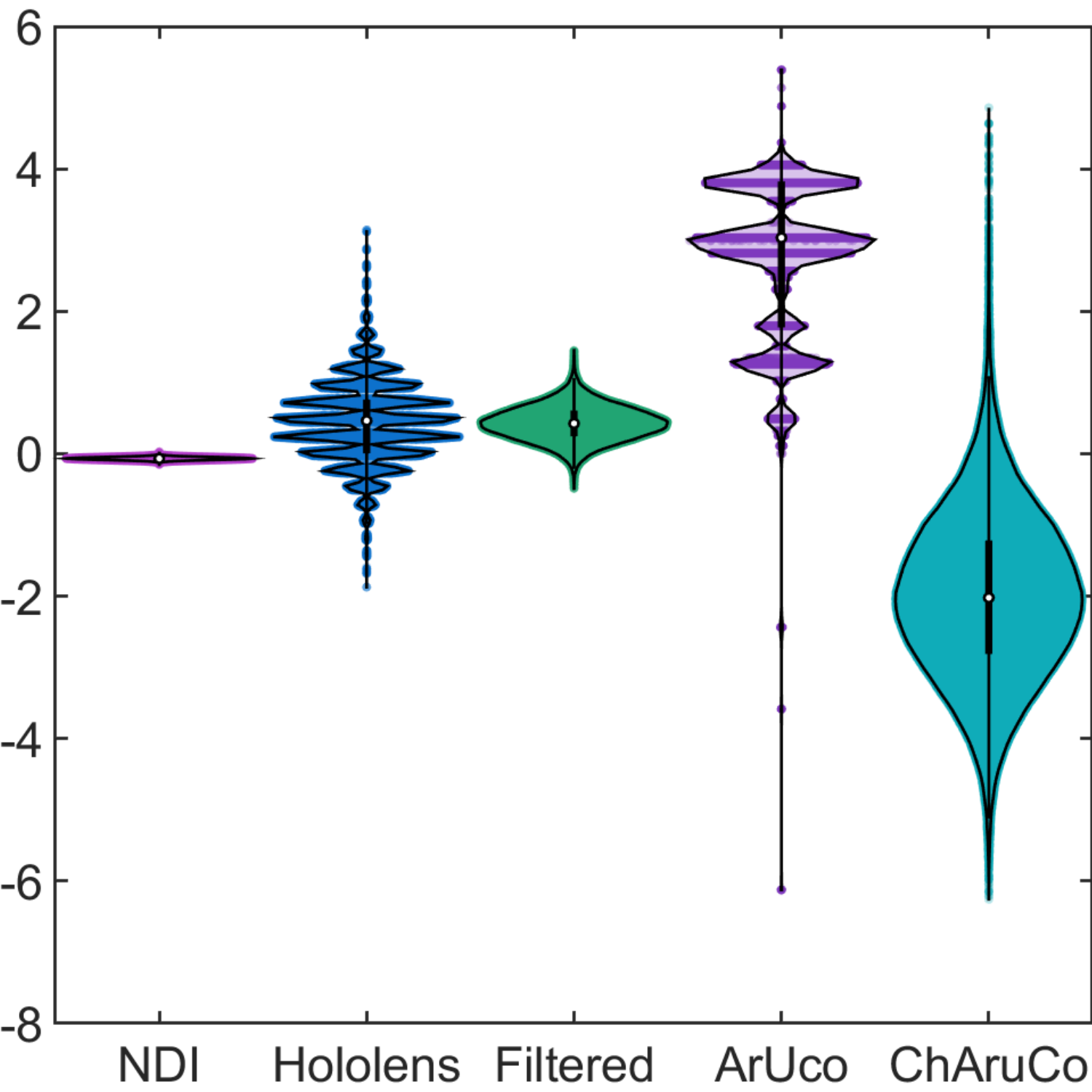}}
			\\
			\scriptsize{(c) 1 mm translation \textit{z-axis}} & & \scriptsize{(d) 20 mm translation \textit{z-axis}} \\
			\cmidrule(l){1-1}\cmidrule(l){3-3}
			\raisebox{-\totalheight}{\includegraphics[page=11,width=0.175\textwidth]{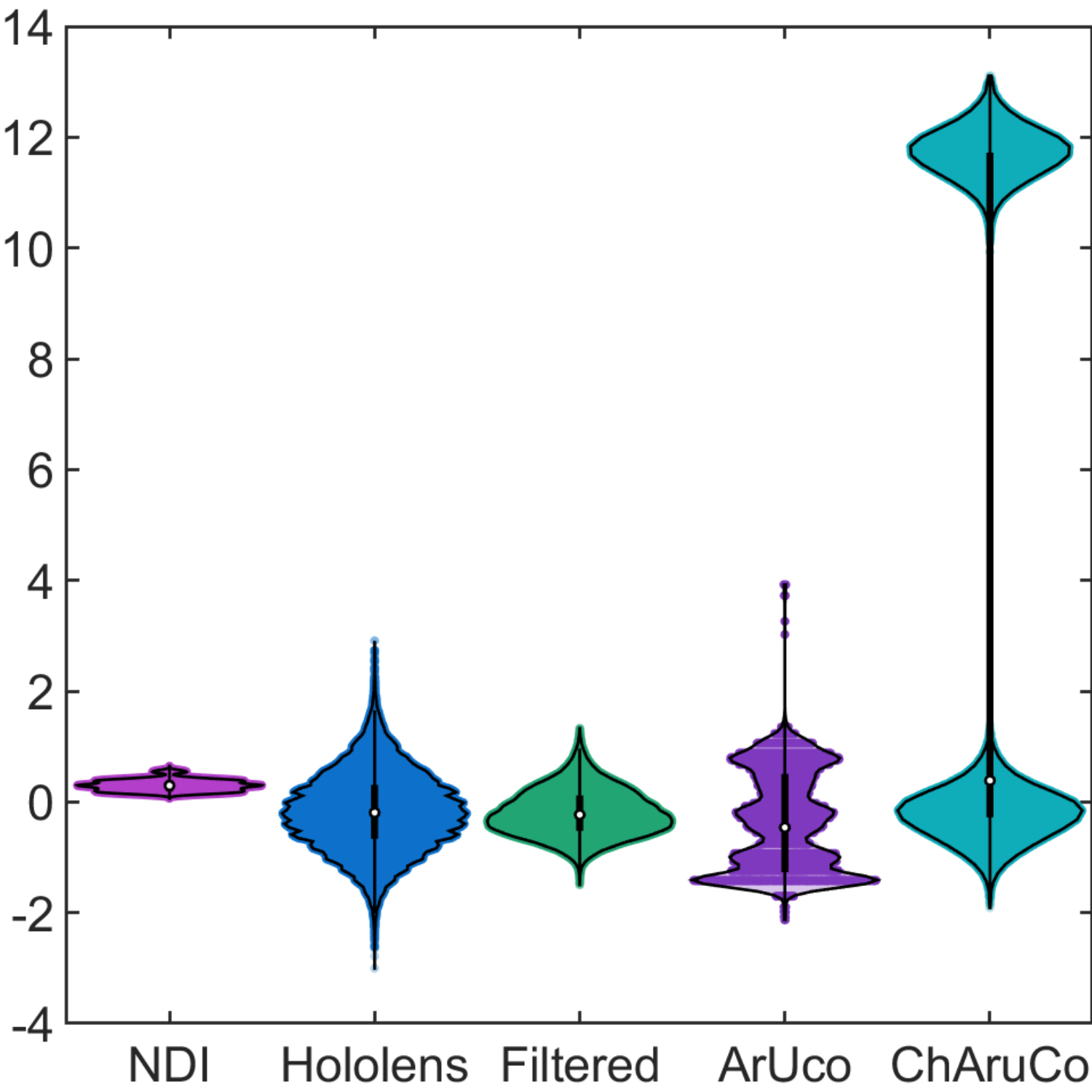}}
			& &
			\raisebox{-\totalheight}{\includegraphics[page=12,width=0.175\textwidth]{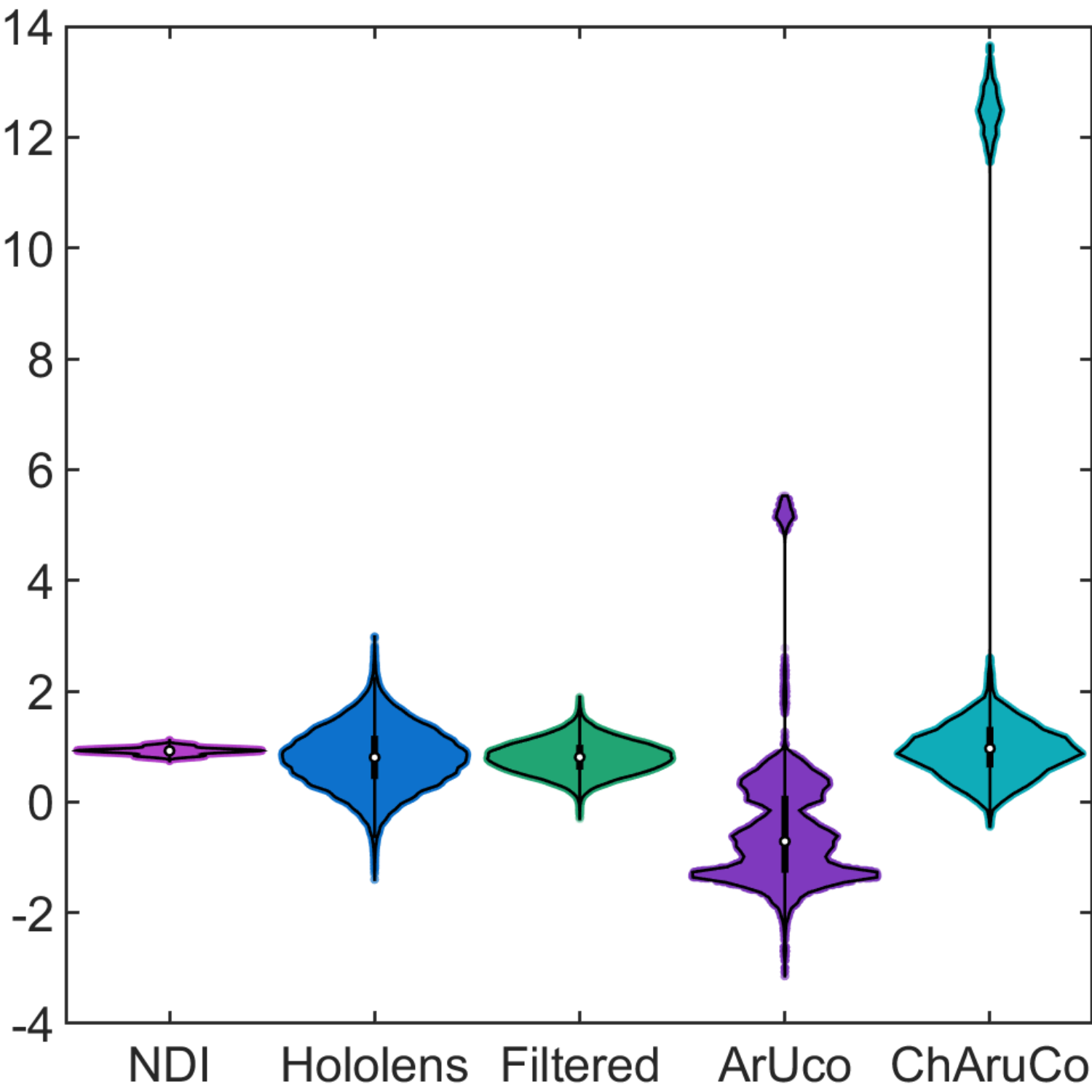}}
			\\
			\scriptsize{(e) $10^\circ$ rotation} & & \scriptsize{(f) $50^\circ$ rotation} \\ 
			\cmidrule(l){1-1}\cmidrule(l){3-3}
		\end{tabular}
		\caption{Experiment results of tool tracking accuracy assessment of our infrared depth camera and Kalman filter-based tracking method, comparing with commercial IR tracking method and visible light marker tracking methods. (a),(c) Detected moving distances when moving optical platform 1 mm along x- and z-axis. (b),(d) Detected moving distances when moving optical platform 20 mm along x- and z-axis. (e),(f) Detected rotation angle when rotating the platform 10 and 50 degrees.}
    \label{fig:Violins}
	\end{center}
\end{figure}

To evaluate the stability and tracking accuracy, we performed a statistical analysis of the results collected.
Considering the non-normal distribution of the data collected, we used Kruskal-Wallis tests with $\alpha=0.05$ to compare the results obtained for position and orientation.
Posterior Bonferroni tests were used to reveal significant differences between the tracking technologies.
The median and inter-quartile range (IQR) of all the experiments are summarized in Table~\ref{tab:Accuracy} and presented in Figure~\ref{fig:Violins}.

Overall, the lowest errors were reported by the NDI tracking system, followed by the proposed method with and without the addition of Kalman filters.
When evaluating the results for 1 mm translations in the \textit{x-axis}, the Kruskal-Wallis test revealed a significant interaction between the multiple tracking technologies ($\chi^{2}(4)=498650.94, p=0$).
A posterior Bonferroni test revealed statistical significance among all the tracking technologies ($p=0$).
The NDI tracking system provided significantly higher accuracy than the other tracking technologies, followed by the proposed method with and without Kalman filters and the ChArUco and ArUco markers (see Figure~\ref{fig:Violins}a).
Regarding the 20 mm translations in the \textit{x-axis}, the Kruskal-Wallis test revealed a significant interaction between the multiple tracking technologies ($\chi^{2}(4)=352721.55, p=0$).
A posterior Bonferroni test showed that the NDI tracking system and the ArUco markers provided significantly higher accuracy than the ChArUco and the proposed method with and without the addition of the Kalman filters ($p=0$).
However, the results presented in Figure~\ref{fig:Violins}b show that the precision provided by the ArUco markers is lower than for all the other tracking technologies.

Regarding position accuracy on the \textit{z-axis}, a Kruskal-Wallis test revealed a significant interaction between the tracking technologies when the markers were translated 1 mm ($\chi^{2}(4)=60421.54, p=0$).
A posterior Bonferroni test revealed statistical significance among all the tracking technologies ($p=0$).
The NDI tracking system proved to be more accurate than the other compared technologies.
The proposed method with and without Kalman filters followed the NDI system, while the ChArUco and ArUco reported the worst scores (Figure~\ref{fig:Violins}c).
When translating the markers 20 mm, our Kruskal-Wallis test revealed a significant interaction between the multiple technologies ($\chi^{2}(4)=809777.32, p=0$).
The posterior Bonferroni test revealed the same behavior then the one observed for the 1 mm translations in this axis (see Figure~\ref{fig:Violins}d).

Furthermore, results from the Kruskal-Wallis tests for the orientation errors revealed a significant interaction between the tracking technologies when rotating the target by $10$ ($\chi^{2}(4)=193849.19, p=0$) and $50$ ($\chi^{2}(4)=377705.61, p=0$) degrees.
In contrast to the translation results, both versions of the proposed method reported lower median errors than the NDI tracking system and the ArUco and ChArUco markers.
However, the precision reported by the NDI tracking system is higher than for the other tracking technologies (see Figures~\ref{fig:Violins}e and \ref{fig:Violins}f).

\begin{table*}[th]
	\caption{Median and IQR accuracy errors reported by the different tracking technologies.}
	\label{tab:Accuracy}
	\scriptsize%
	\centering%
	\begin{tabular}{c*{15}{c}}
		\toprule
		\multicolumn{1}{c}{}& 
		\multicolumn{1}{c}{}& 
		\multicolumn{4}{c}{Translation (\textit{x-axis})} &
		\multicolumn{1}{c}{}& 
		\multicolumn{4}{c}{Translation (\textit{z-axis})} &
		\multicolumn{1}{c}{}& 
		\multicolumn{4}{c}{Rotation}\\
		\cmidrule{3-6} \cmidrule{8-11} \cmidrule{13-16}
		\multicolumn{1}{c}{Tracking}
		&
		& \multicolumn{2}{c}{1 mm} & \multicolumn{2}{c}{20 mm} 
		&
		& \multicolumn{2}{c}{1 mm} & \multicolumn{2}{c}{20 mm}
		&
		& \multicolumn{2}{c}{10$^\circ$} & \multicolumn{2}{c}{50$^\circ$}\\
		\multicolumn{1}{c}{Technology}
		&
		& Median & \multicolumn{1}{c}{IQR} & Median & \multicolumn{1}{c}{IQR} 
		&
		& Median & \multicolumn{1}{c}{IQR} & Median & \multicolumn{1}{c}{IQR}
		&
		& Median & \multicolumn{1}{c}{IQR} & Median & \multicolumn{1}{c}{IQR}\\
		\midrule
		\multicolumn{1}{c}{ArUco}
		&
		& 1.0566 & \multicolumn{1}{c}{2.6093} & -0.020 & 0.267 
		&
		& 1.0544 & \multicolumn{1}{c}{3.0490} & 3.032 & 2.001
		&
		& -0.461 & \multicolumn{1}{c}{1.717} & -0.713 & 1.339
		\\
		\multicolumn{1}{c}{ChArUco}
		&
		& 0.3047 & \multicolumn{1}{c}{0.7078} & 0.047 & 0.195 
		&
		& 0.4215 & \multicolumn{1}{c}{1.633} & -0.022 & 1.553
		&
		& 0.386 & \multicolumn{1}{c}{11.941} & 0.964 & 0.680
		\\
		\multicolumn{1}{c}{NDI}
		&
		& -0.0007 & \multicolumn{1}{c}{0.0172} & -0.028 & 0.014 
		&
		& 0.008 & \multicolumn{1}{c}{0.027} & -0.067 & 0.026
		&
		& 0.294 & \multicolumn{1}{c}{0.151} & 0.922 & 0.075
		\\
		\multicolumn{1}{c}{Ours}
		&
		& 0.1223 & \multicolumn{1}{c}{0.2226} & -0.089 & 0.083
		&
		& 0.058 & \multicolumn{1}{c}{0.661} & 0.464 & 0.714
		&
		& -0.194 & \multicolumn{1}{c}{0.917} & 0.804 & 0.728
		\\
		\multicolumn{1}{c}{Ours + Kalman}
		&
		& 0.0764 & \multicolumn{1}{c}{0.1332} & -0.092 & 0.063 
		&
		& -0.010 & \multicolumn{1}{c}{0.289} & 0.424 & 0.320
		&
		& -0.231 & \multicolumn{1}{c}{0.583} & 0.807 & 0.395
		\\
		\bottomrule
	\end{tabular}%
\end{table*}

\subsubsection*{Workspace Definition}
In addition to the tracking accuracy, a further experiment investigated the role that the FoV plays over the proposed method's accuracy.
To achieve more considerable displacement capabilities than the one used in the previous experiment, we moved our tracking tool along the \textit{x-} and \textit{z-axis} using a different linear stage.
For this portion of the experiment, the NDI tracking system was used as the ground truth to evaluate the tracking accuracy.
The passive tool was incrementally moved along the linear stage.
After the movement of the linear stage was completed, the data corresponding to the tool's pose was collected using the NDI tracking system and the AHAT camera of the HoloLens.
A total number of 50 values were collected for every position.
This process was repeated from the nearest to the furthest detection distances for the \textit{z-axis} and from the center to the lateral margins of the \textit{x-axis} that could be detected using the AHAT camera.
The mean values of the data at every position were considered the real pose values and used to estimate the moving direction of the tool.
The estimated displacement along the moving platform can then be calculated by projecting the tracked tool's position to the moving direction and comparing it with its initial position.

\begin{figure}[!t]
    \centering
        \subfloat[\label{fig:GlobalPrecision_A}]{\includegraphics[page=1,width=0.475\columnwidth]{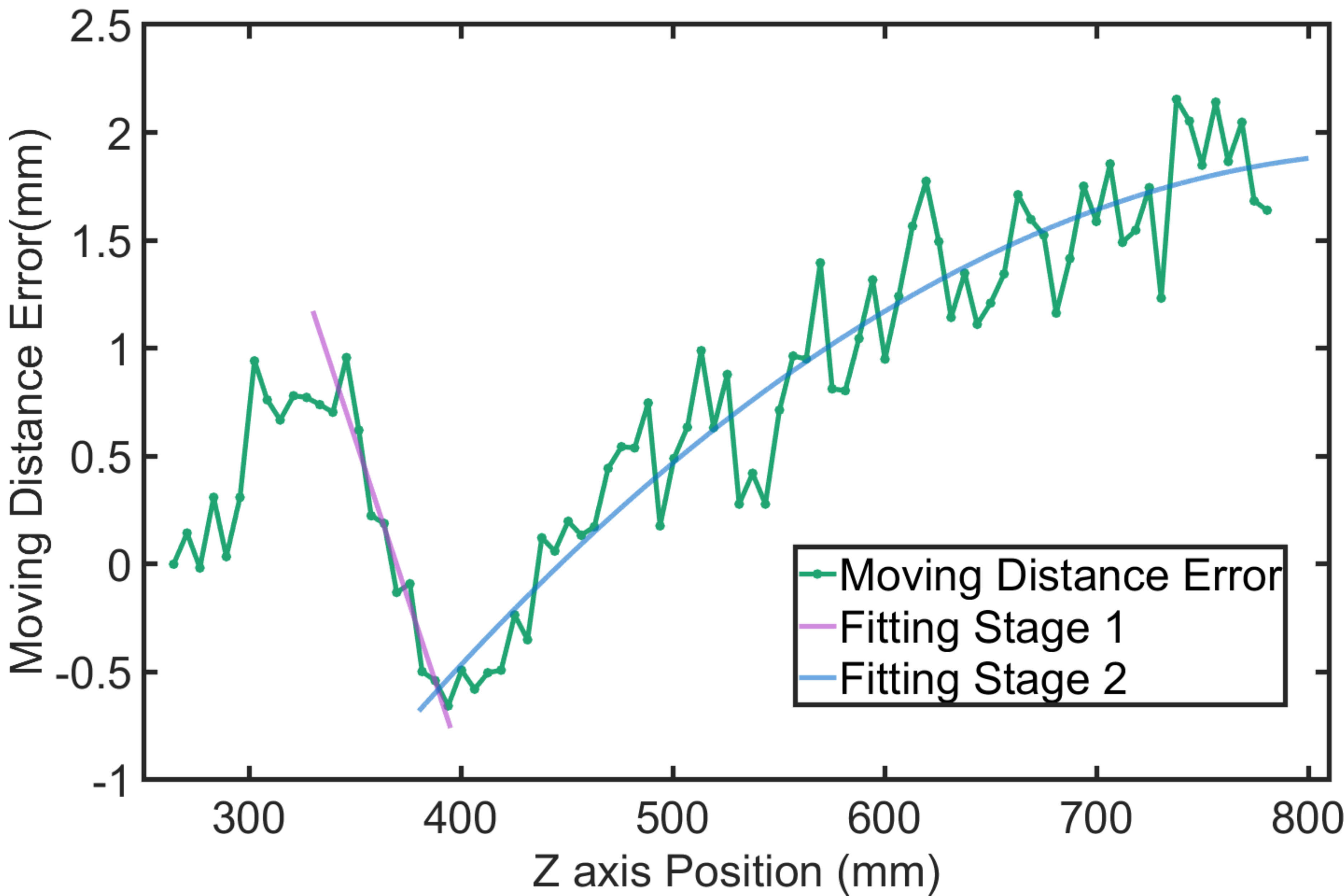}}\hfill
        \subfloat[\label{fig:GlobalPrecision_B}]{\includegraphics[page=2,width=0.475\columnwidth]{Figures/GlobalPrecision_v4.pdf}}
    \caption{The global tracking accuracy of the AHAT tracking method compared to NDI Polaris Spectra. \protect\subref{fig:GlobalPrecision_A} Moving distance detection error along the \textit{z-axis}. \protect\subref{fig:GlobalPrecision_B} Moving distance detection error along the \textit{x-axis} at 500mm depth.}
    \label{fig:GlobalPrecision}
\end{figure}

The detection error observed during tool displacement using our system is shown in Figure~\ref{fig:GlobalPrecision} as a function of the observed depth.
Such results show that our system can steadily trace the passive tool within depths of $250$ and $750 mm$ (Figure~\ref{fig:GlobalPrecision_A}) and within radial distances of $509 mm$ when the tool is placed at a maximum distance of $510 mm$ (Figure~\ref{fig:GlobalPrecision_B}).
Using $FoV=2\cdot arctan(x_{max}/d)$, it can be shown that these values are equivalent to an FoV of $89.9^\circ$.
Interestingly, the detection error observed for the \textit{z-axis} depicts a different behavior before and after $400 mm$.
In addition, the absolute difference between the error observed at $400$ and $800 mm$ results in a moving distance error of $2.5 mm$, or a distortion of $0.625\%$.
For the radial direction, a smaller moving distance error can be observed.
The absolute difference observed between 0 and $500 mm$ translations depicts an error of $\approx 0.75 mm$, equivalent to a distortion of $0.15\%$.

\subsection{Runtime and Latency}
To investigate the runtime capabilities of our system, we placed different quantities of passive tools within the FoV of the AHAT camera.
The sensor data collected by the AHAT was recorded and used for the offline runtime test.
In this way, the frame rate of the algorithm during the test would not be limited by the sensor acquisition frequency.
In addition, we collected and evaluated the runtime for detecting the ArUco and ChAruCo markers using an RGB camera with a resolution of $1280\times 1024$ pixels.
We used the average runtime of 10,000 frames on a Xeon(R) E5-2623 v3 CPU for this portion of the experiments.

\begin{figure}[!t]
    \centering
        \subfloat[\label{fig:Framerate}]{\includegraphics[page=1,width=0.475\columnwidth]{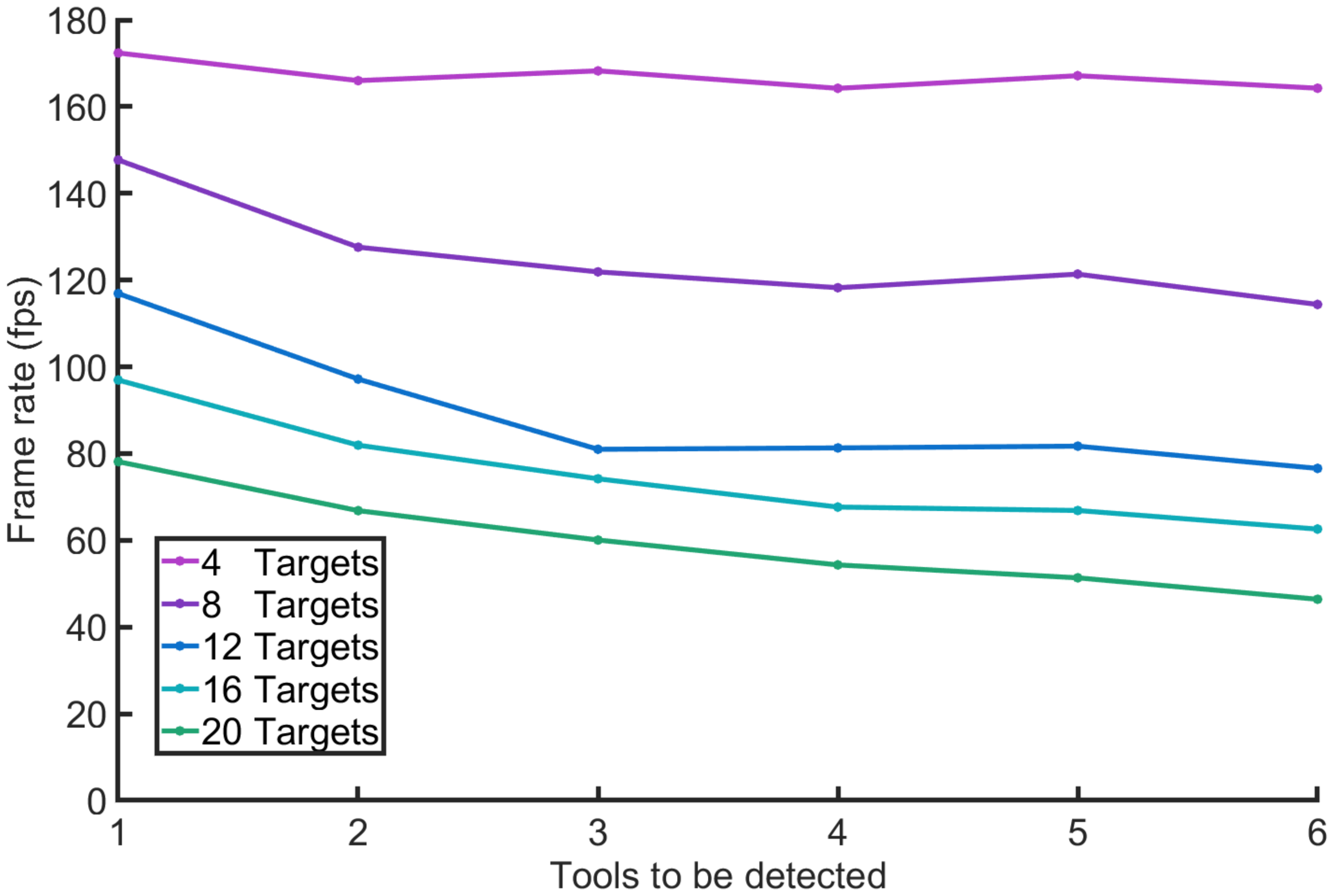}}\hfill
        \subfloat[\label{fig:Latency}]{\includegraphics[page=2,width=0.475\columnwidth]{Figures/FrameRate_v4.pdf}}
    \caption{Frame rate and latency results of the proposed algorithm using the AHAT camera. \protect\subref{fig:Framerate} Tool detection frame rate under different conditions. Every series depicts several passive tools containing four retro-reflective markers detected using the AHAT camera. Different tools are loaded to evaluate the detection frame rate when multiple markers are observed. If more tools are defined than the number of markers observed, the system will indicate that the respective tool is not detected. \protect\subref{fig:Latency} Movement detection of synchronously acquired localization data from the NDI tracking system and the proposed method. The target is kept in motion back and forth using a linear stage.}
    \label{fig:FramerateAndLatency}
\end{figure}

This experiment revealed a runtime of 28.4 ms (35.2 Hz) for the detection of the ArUco and 30.14 ms (33.2 Hz) for the ChArUco makers.
The runtime results when using the AHAT camera proved to be influenced by both the number of retro-reflective markers detected in the scene and the number of loaded tools for detection.
These results are depicted in Figure~\ref{fig:Framerate}.
When only one passive tool is expected to be detected, the runtime of our system is 5.80 ms (172.37 Hz).
However, when the passive markers corresponding to five different tools are detected and their five corresponding definitions are loaded, the runtime is 19.48 ms (51.34 Hz).

The latency of our system was then compared using the NDI tracking system as a reference.
For this experiment, we moved back and forth a passive tool attached to a linear stage using a period $T_{mov}$ of $\approx 15$ seconds.
The localization data from the NDI and our system were collected simultaneously.
The detected moving distance along the linear stage for both systems can be expressed as $N(t)$ and $H(t)$ with $t\in [0,T)$ for the NDI and HoloLens, respectively (see Figure~\ref{fig:Latency}).
The moving signal of the HoloLens is first adjusted to match the amplitude of the NDI system using:
\begin{equation}
    \begin{matrix}
        \scriptsize{H'(t) = H(t) + \frac{\delta(N(t))-\delta(H(t))}{2}} \\ \\
        \delta (x) = max(x) + min(x)
    \end{matrix}
\end{equation}

The following step computed the delay between our tracking system and the NDI system using:
\begin{equation}
    \delta_H=\mathop{argmin}\limits_{0<\delta<T_{mov}} \int_{\delta}^T \frac{1}{T-\delta} (N(t)-H'(t-\delta))^2dt
\end{equation}
\noindent where $\delta_H$ depicts the time difference in seconds between the signals $H'(t)$ and $N(t)$.

This experiment showed that the proposed method presents a time delay of 103.23 ms compared to the NDI tracking system.

\section{Use Case}\label{sec:UseCase}
\begin{figure}[!t]
    \centering
    \includegraphics[page=3,width=0.9\columnwidth]{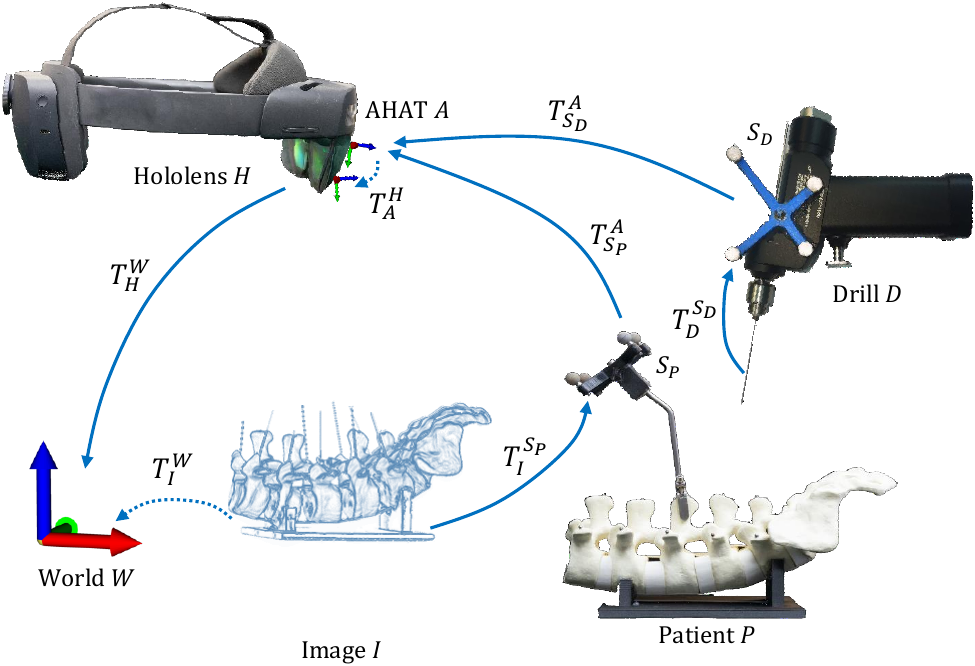}
    \caption{The spatial relationship between different components is required to provide surgical navigation using the proposed framework.}
    \label{fig:Spatial_Relationship}
\end{figure}

To highlight the relevance of the proposed framework for surgical applications, we introduced a use case in which we provided visual navigation using AR HMDs in orthopedic surgical procedures for the placement of pedicle screws.
Two different tools containing retro-reflective markers were attached to a phantom spine model ($S_P$) and a surgical drill ($S_D$) to enable visual navigation.
We used an O-arm imaging system and acquired a pre-operative computed tomography (CT) scan from the spine model.
This data was used to plan the trajectories that the surgeons must follow during the placement of the pedicle screws.
In addition, we added four metallic balls (BBs) to the phantom spine model that were visible in the CT and by direct observation.
This action allowed us to determine the spatial relationship between the tool attached to the spine model ($S_P$) and the image space ($I$).
The spatial relationship between the different coordinate systems involved in this use case is depicted in Figure~\ref{fig:Spatial_Relationship}.

After successful tracking of the retro-reflective tools, the pose of the phantom model in world coordinates can be computed using the tool pose $T_{S_P}^A$, estimated with the proposed framework, and the HMD's self-localization data $T_{H}^W$ as follows:
\begin{equation}
    T_I^W=T_H^W \cdot T_A^H \cdot T_{S_P}^A \cdot T_I^{S_P}
\end{equation}
Likewise, the pose of the surgical drill in world coordinates can be acquired using the tool pose $T_{S_D}^A$ utilizing:
\begin{equation}
    T_D^W=T_H^W \cdot T_A^H \cdot T_{S_D}^A \cdot T_{D}^{S_D}
\end{equation}

This procedure requires calculating the transformation matrix from the AHAT camera to HoloLens view space ($T_A^H$).
For this purpose, we designed a 3D-printed structure composed of a retro-reflective tool, $S_M$, and four BBs to register the AHAT and view spaces (see Figure~\ref{fig:Teaser_A}).
To calculate the position of the real BBs in the tracker space, we performed a pivot calibration.
The spatial information extracted from the real BBs during the pivot calibration served to generate a set of virtual replicas that were aligned to their real counterparts.
A computer keyboard enabled the user to control the 6 DoF of the virtual objects in the HoloLens world space ($T_M^W$).
After proper alignment of the real and virtual objects, the spatial relationship between the AHAT camera and the view space ($T_A^H$) was computed as follows:
\begin{equation}
    T_A^H=(T_H^W)^{-1}\cdot T_M^W\cdot (T_{S_M}^A)^{-1}
    \label{eq:Calibration}
\end{equation}

\subsection{Surgical Navigation}
To further evaluate the performance of our surgical navigation system, we designed a total of three phantom models to replicate traditional k-wire placement procedures (Figure~\ref{fig:Teaser_B}).
These models were composed of a 3D printed base containing six cylindrical shapes with conical tips covered using silica gel (see Figure~\ref{fig:Teaser_B}).
After constructing the phantom models, we collected CT volumes from them to generate their virtual replicas.
An experienced surgeon later used the CT volumes to plan multiple trajectory paths aiming at the tips of the individual cones.
A total of two different drilling paths were planned for every cone, leading to a total of 12 trajectories per phantom model.
These trajectories defined the optimal path in which a k-wire should be inserted and were overlayed on the phantom using the HoloLens.
In addition to the trajectories, virtual indicators in the form of concentric circles were presented over the surface of the phantom.
The virtual prompts indicated the entry point that would lead to the optimal trajectory in the physical model (Figure~\ref{fig:Teaser_C}).
A total number of three participants, including two experienced surgeons and a biomedical researcher with experience in mixed reality, took part in this portion of the study.
Among the surgeons, one reported comprehensive experience using mixed reality, while the other was unfamiliar with this technology.

Every participant performed an eye-tracking calibration procedure before the start of the experiment.
After calibration, participants were asked to complete the registration procedure described in Equation~\ref{eq:Calibration} and depicted in Figure~\ref{fig:Teaser_A}.
Once the virtual and real models were registered and the spatial relationship between the AHAT camera and the view space was computed, the pre-planned trajectories were presented to the participants using the AR HMD.
Before collecting data for evaluation, every participant was allowed to perform multiple k-wire insertions in one of the three models to get familiar with the system.
After this step, the remaining models containing 24 drilling paths were used for formal testing.
Among these trajectories, three were not considered for evaluation because the construction of the model and the retro-reflective markers' placement interfered with the planned trajectory's achievement.
Therefore, we considered 21 drilling trajectories per participant for evaluation.
To measure the accuracy achieved by the study participants, a registration step between the pre- and post-operative imaging was performed.
This step involved the acquisition of CT volumes from the models and allowed comparing the differences between the planned and real trajectories.
The translation and angular errors between the optimal and real trajectories were used as metrics for the performance evaluation.

To further evaluate the results obtained in this portion of the study, we performed a statistical analysis using the data collected for the translation and angular errors.
The results obtained for position and orientation were compared using ANOVA tests with $\alpha=0.05$ .
Posterior Tukey-Kramer tests revealed significant differences between the participants' performance.

\begin{figure}[!t]
    \centering
        \subfloat[\label{fig:ModelTest_Translation}]{\includegraphics[page=1,width=0.475\columnwidth]{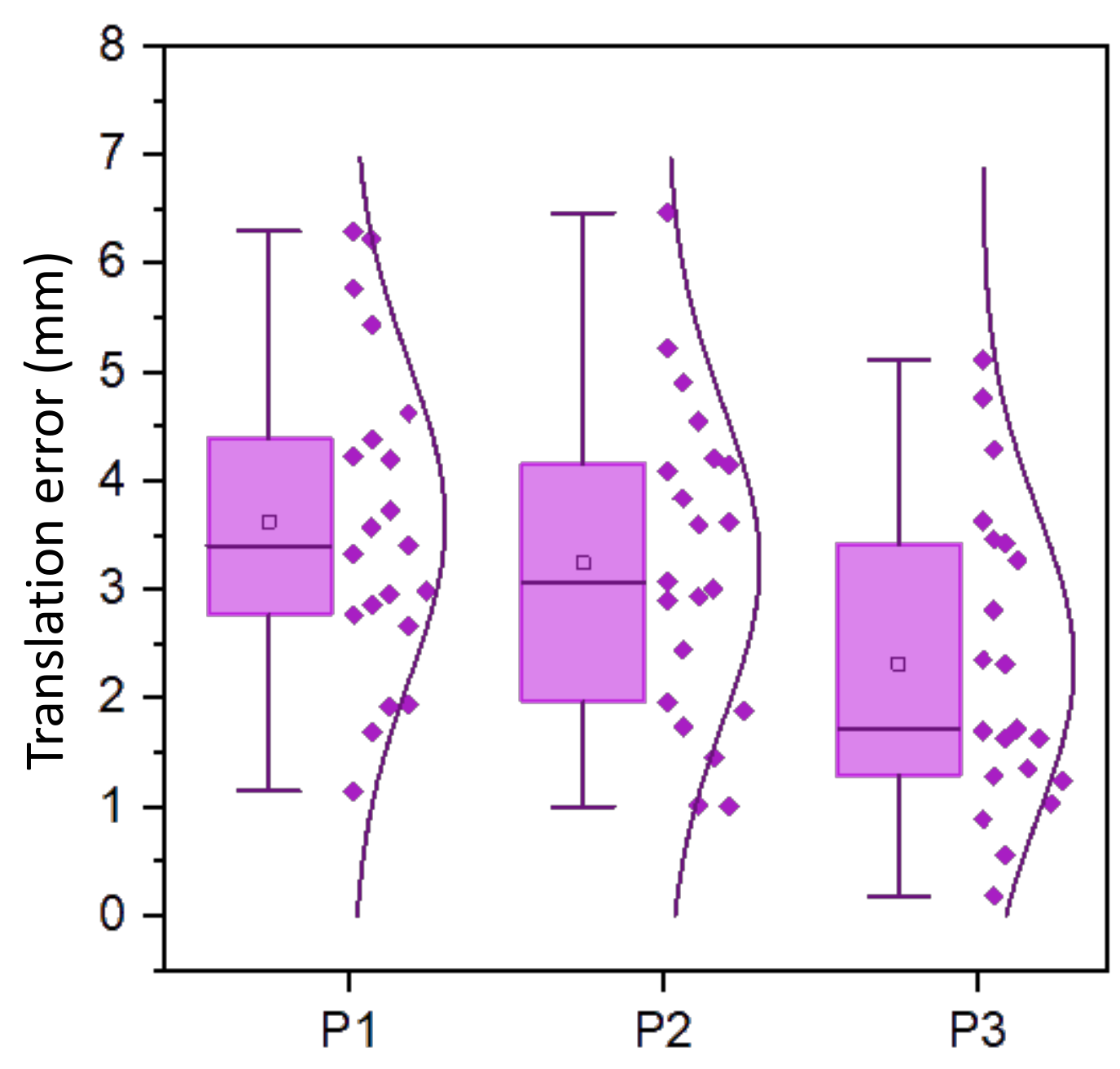}}\hfill
        \subfloat[\label{fig:ModelTest_Orientation}]{\includegraphics[page=2,width=0.475\columnwidth]{Figures/UseCase.pdf}}
    \caption{\protect\subref{fig:ModelTest_Translation} Translation and \protect\subref{fig:ModelTest_Orientation} orientation errors for k-wire insertion using AR visual guidance and the proposed algorithm for tool tracking. P1: Surgeon with reported experience using AR-guidance systems. P2: Surgeon without reported experience using AR. P3: Researcher with reported experience using AR.}
    \label{fig:ModelTest}
\end{figure}

Results from this experiment, summarized in Figure~\ref{fig:ModelTest}, showed a significant interaction between the accuracy in position achieved by the users.
Our ANOVA test revealed significant interaction between the participants for alignment ($F(2)=4.65, p=0.0133$).
The biomedical researcher achieved better alignment scores ($Mn=2.32mm, SD=1.40, p=0.0119$) when compared to the experienced surgeon ($Mn=3.64mm, SD=1.47$).
However, no significant differences were found between the biomedical researcher and the surgeon unfamiliar with AR-based navigation systems ($Mn=3.25mm, SD=1.45$) or between the surgeons.
In addition, an ANOVA test for the rotation scores did not reveal a significant interaction between the participants ($F(2)=2.96, p=0.0593$).
However, as depicted in Figure~\ref{fig:ModelTest_Orientation}, the experienced surgeon achieved better alignment ($Mn=3.23^\circ, SD=1.95$) than the other surgeon ($Mn=4.95^\circ, SD=2.66$) and the biomedical researcher ($Mn=4.36^\circ, SD=2.27$).

These results show no significant differences in translation or orientation between the surgeons when using the system.
This is particularly interesting as the surgeon with experience using AR systems also reported an approximate experience of 10 years working with navigation systems for k-wire injection procedures, while the surgeon that did not report experience with AR-based navigation systems also reported an approximate experience of 1 year in the field.
The scores reported by the participants of the study are comparable to the accuracy values reported by other AR-based navigation systems for pedicle screw placement \cite{liebmann2019pedicle,muller2020augmented,spirig2021augmented}.

Although this experiment shows that the proposed methods provide comparable tracking accuracy and stability to successfully performing the insertion of the k-wires, the latency observed between the data processing using the workstation and the visualization of the optimal trajectories using the HMD seems to contribute to the observation of uncertainty during the alignment and localization of the targets.
Furthermore, the elastic and smooth surface of the silicone gel used to create the test models replicates the challenges brought by percutaneous surgeries where few landmarks are visible, and the working surface is soft and deformable.
These properties increase the difficulty of performing the drilling task and the accurate insertion of the k-wire. 
However, the scores reported by the users demonstrate the performance of the proposed system when these challenges exist.

\section{Discussion}\label{sec:Discussion}
This work introduces a framework to track retro-reflective markers using the built-in cameras of commercially available AR HMDs.
Results regarding the error distribution of depth detection demonstrate that the error of the AHAT depth data can be modeled using a normal distribution.
This normal distribution increases its standard error as a function of the detected depth (see Figure~\ref{fig:Gauss_Test_Scene}). 
Results from the test of tracking accuracy demonstrate that the integration of Kalman filters contributes to the reduction of the IQR, as depicted in Table~\ref{tab:Accuracy}.
Therefore, they contribute to improve the precision of the methods proposed.
As a result of the integration of the Kalman filters, this framework showed to be capable of achieving a tracking accuracy of $0.09 \pm 0.06\ mm$ on lateral translation, $0.42 \pm 0.32\ mm$ on longitudinal translation, and $0.80\pm 0.39^\circ$ on rotation around the vertical axis (see Figure~\ref{fig:precision} and Table~\ref{tab:Accuracy}). 
These results are more accurate and precise than those achievable using traditional feature-based tracking algorithms such as the ArUco and ChArUco markers.
Although it can be expected that feature-based tracking algorithms provide lower accuracy and precision, they are frequently used to enable the optical tracking of tools in AR applications.
Therefore, they were considered during our study.
Of note, our statistical analysis did not reveal statistical significance when comparing the tracking results of the NDI tracking system and the ArUco markers for 20 mm movements on the \textit{x-axis}.
However, when observing the distribution of the tracking results of the ArUco markers, the precision provided by this type of marker is worse than the one observed when using the NDI tracking system and the two variants of the proposed framework.

In terms of applicability, the use case for the placement of k-wires during orthopedic interventions demonstrates its potential use in medical environments during the performance of surgical procedures \cite{liebmann2019pedicle,muller2020augmented,spirig2021augmented}.
Although the biomedical researcher reported significantly more accurate scores for translation during the placement of the k-wires, no significant differences in translation and orientation were found between the surgeons of the use case.
These results may suggest that using this technology can help novice users reach similar results to those achieved by experienced surgeons, even with little time to get familiar with the system.
However, further studies would need to be conducted in this regard.

In addition, the proposed framework was demonstrated to be capable of tracking individual tools containing up to 4 retro-reflective markers at a frame rate of 172 Hz.
This tracking frame rate remains over 50 Hz even when five different tools containing multiple retro-reflective markers are detected simultaneously.
These reported speeds are faster than that of traditional feature-based tracking techniques such as ArUco (35 Hz) and ChArUco (33.2 Hz) markers.
This frame rate is even higher than the frame rate of the AHAT camera (45 Hz).
Therefore, allowing for tracking multiple targets before having a new sensor frame available.

In contrast to existing works, the proposed framework only uses the ToF camera of the HoloLens 2 without the addition of external components.
Moreover, when compared to methods that use the grayscale environmental cameras of the headset for tool tracking, several benefits support using the depth sensor when hand distance tracking is needed.
In this case, the tracking distance from 250 mm to 750 mm provided by the AHAT sensor nicely satisfies the specific application's needs.
At the same time, the LF and RF cameras have short baseline distances and focal lengths, restricting their performance at near distances \cite{Gsaxner:2021:IO_HoloLend_Navigation}.
In addition, using the AHAT camera provides an FoV of $\approx90^\circ$, while the overlap of the LF and RF cameras is smaller than $60^\circ$ in width. 
Moreover, the proposed algorithm does not require the addition of external infrared light to enhance the visibility of the passive markers.
Therefore, mitigating the possibility of interfering with the functionalities of the HMD that rely on the use of this type of light, including environmental reconstruction and hand tracking.

\subsection*{Limitations}
The utility of the HMD's depth camera for tracking and detecting retro-reflective markers presented in this work led to promising results in terms of precision and stability.
However, certain limitations associated with the proposed framework exist.
In this regard, compared to commercially available tracking systems, the accumulated error observed using our method increases as a function of the depth in which the tools are tracked.
However, this accumulated error remains less than 1 mm at distances between 300 to 600 mm.
In the future, this limitation caused by the precision achieved by the built-in depth sensor of the HMD could be addressed by modeling the error and designing an error compensation algorithm.
Another limitation of the proposed methods is the latency observed when presenting the virtual content to the user after the achievement of tool tracking.
This issue currently represents the most significant challenge in providing visual information that could enable stable and reliable navigation.
As the proposed framework relies on the quality of the network to transfer the data collected, the incorrect synchronization between the tool tracking result $T_{S_P}^A$ and the HMD's self-localization data $T_H^W$ could lead to the observation of inconsistencies in the content displayed.
These differences would be more noticeable when using wireless networks, where the sensor data transfer delay is longer and the frame rate is lower ($<12 fps$).
More importantly, the results presented in this work model the properties of the AR HMD used during the experiments.
Additional studies must be conducted to investigate if the results are consistent for multiple devices of the same type.
\section{Conclusion}\label{sec:Conclusion}
This paper proposes a framework that uses the built-in cameras of commercially available AR headsets to enable the accurate tracking of passive retro-reflective markers.
Such a framework enables tracking these markers without integrating any additional electronics into the headset and is capable of simultaneous tracking of multiple tools. 
The proposed method showed a tracking accuracy of approximately 0.1 mm for translations on the lateral axis and approximately 0.5 mm for translations on the depth axis.
The results also show that the proposed method can track retro-reflective markers with an accuracy of less than $1^\circ$ for rotations.
Finally, we demonstrated the early feasibility of the proposed framework for k-wire insertion as performed in orthopedic procedures.

\ifCLASSOPTIONcompsoc
\else
\fi


\ifCLASSOPTIONcaptionsoff
  \newpage
\fi
\bibliographystyle{IEEEtran}
\bibliography{Bibliography.bib}

%

\begin{IEEEbiography}[{\includegraphics[width=1in,height=1.25in,clip,keepaspectratio]{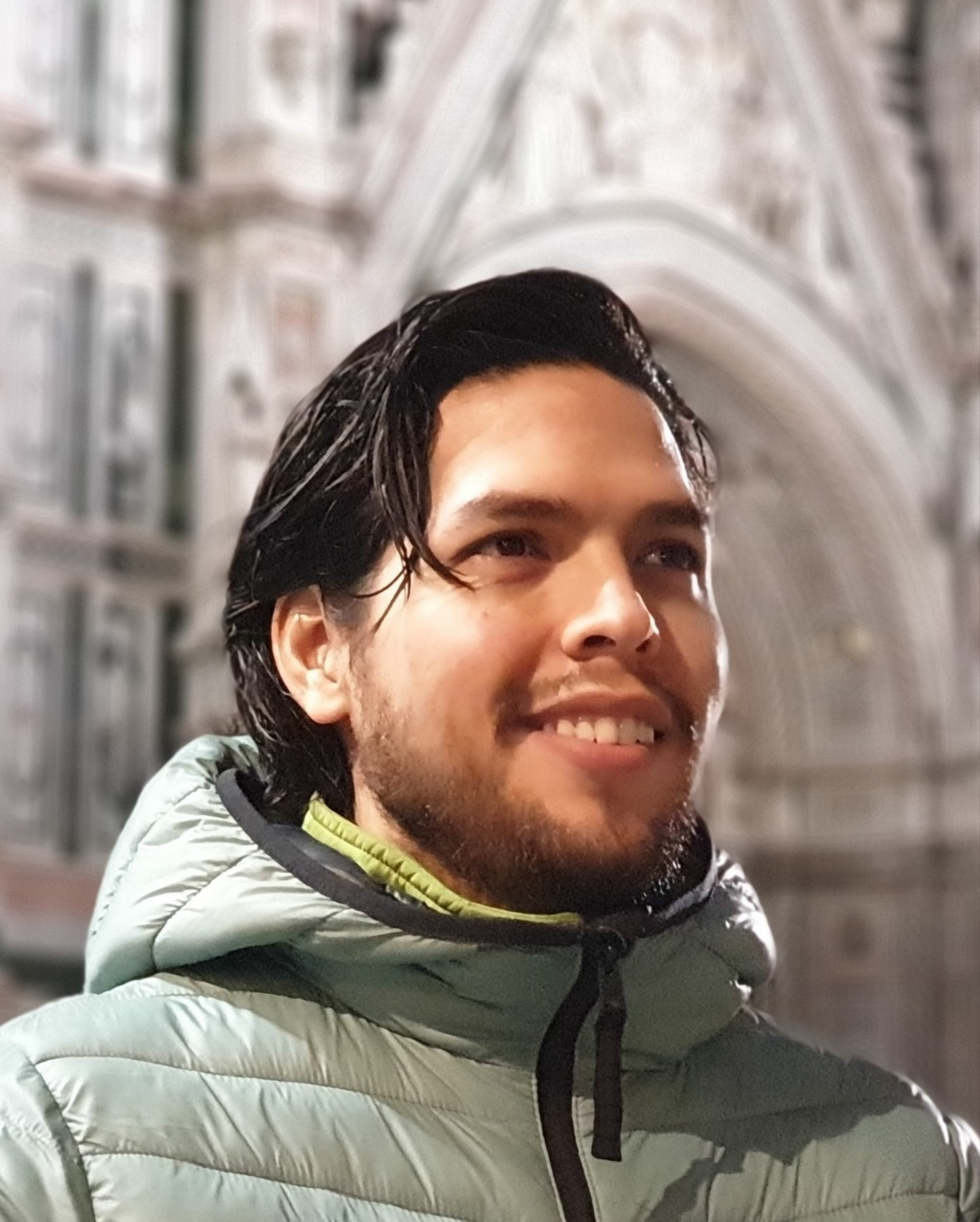}}]{Alejandro Martin-Gomez}
is a Postdoctoral Fellow at the Laboratory for Computational Sensing and Robotics, Johns Hopkins University. Prior to his Ph.D. at the Chair for Computer Aided Medical Procedures and Augmented Reality (CAMP), he earned his M.Sc. degree in Electronic Engineering from the Autonomous University of San Luis Potosi, Mexico, and B.Sc. degree in Electronic Engineering from the Technical Institute of Aguascalientes, Mexico. 
His main research interests include the improvement of visual perception for augmented reality and its applications in interventional medicine.
\end{IEEEbiography}

\begin{IEEEbiography}[{\includegraphics[width=1in,height=1.25in,clip,keepaspectratio]{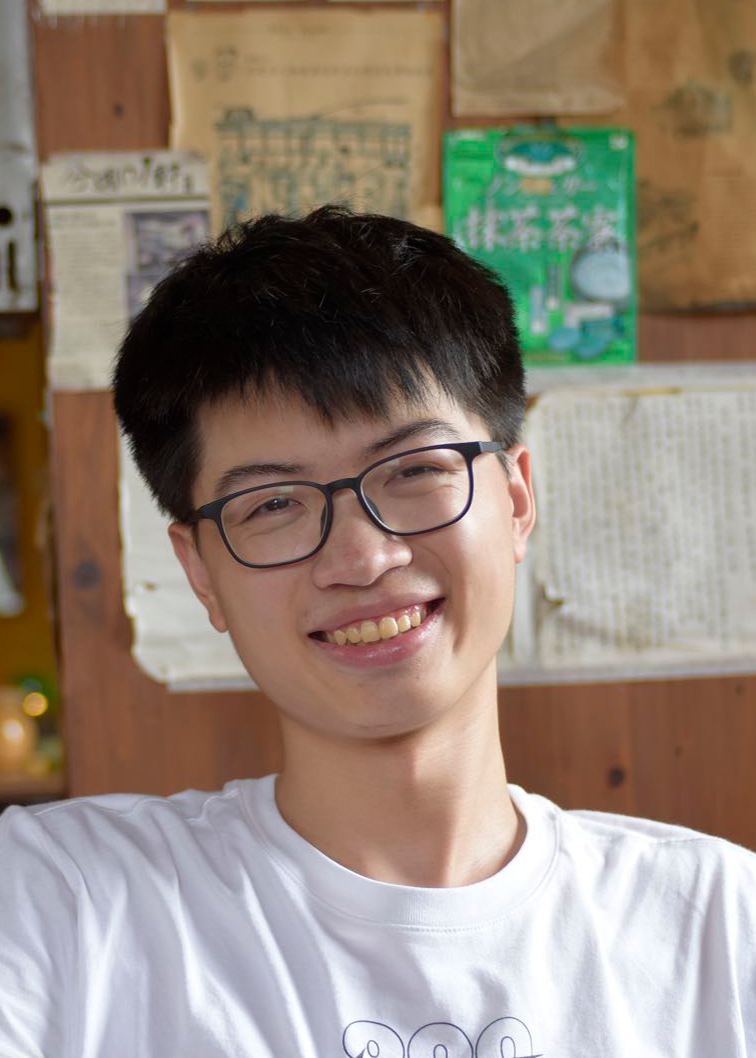}}]{Haowei Li}
is an undergraduate student in the Department of Biomedical Engineering, Tsinghua University. He has been a member of Prof. Guangzhi Wang's lab since junior years. His current interests include computer aided surgery and augmented reality.
\end{IEEEbiography}

\begin{IEEEbiography}[{\includegraphics[width=1in,height=1.25in,clip,keepaspectratio]{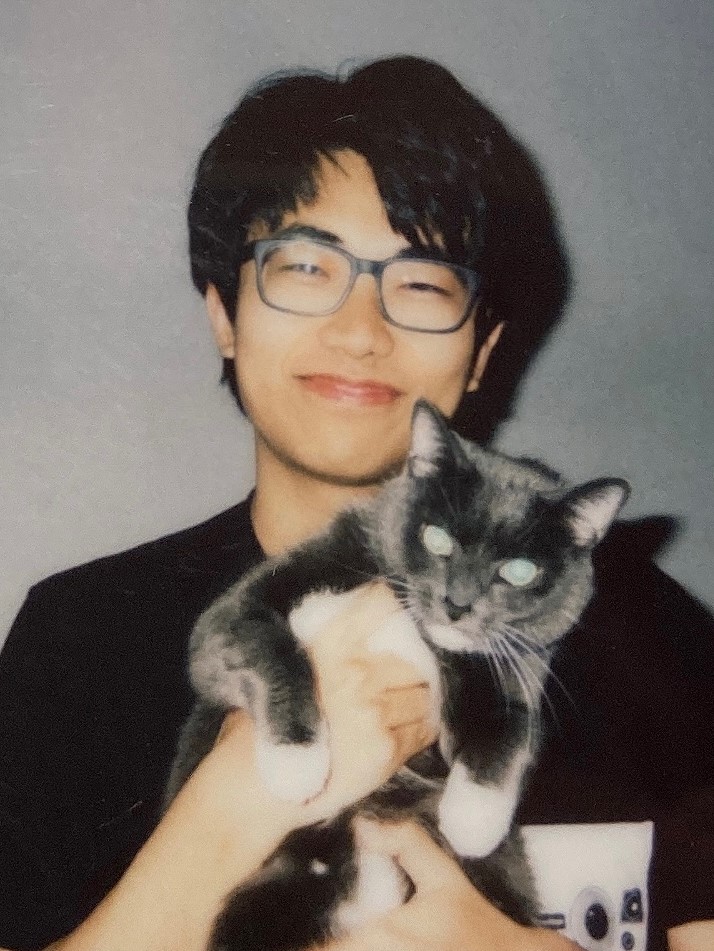}}]{Tianyu Song}
is a Ph.D. candidate of Computer Science at the Chair for Computer Aided Medical Procedures and Augmented Reality (CAMP) of the Technical University of Munich. He received the B.Sc. degree in Theoretical and Applied Mechanics from Sun Yat-sen University, China and B.Sc. degree in Mechanical Engineering from Purdue University, USA in 2017. He earned his M.Sc. in Robotics at Johns Hopkins University, USA in 2019. After graduation, he worked at the Applied Research team at Verb Surgical/ Johnson\&Johnson, USA for a year. His current interests include augmented reality and computer aided surgery.
\end{IEEEbiography}

\begin{IEEEbiography}[{\includegraphics[width=1in,height=1.25in,clip,keepaspectratio]{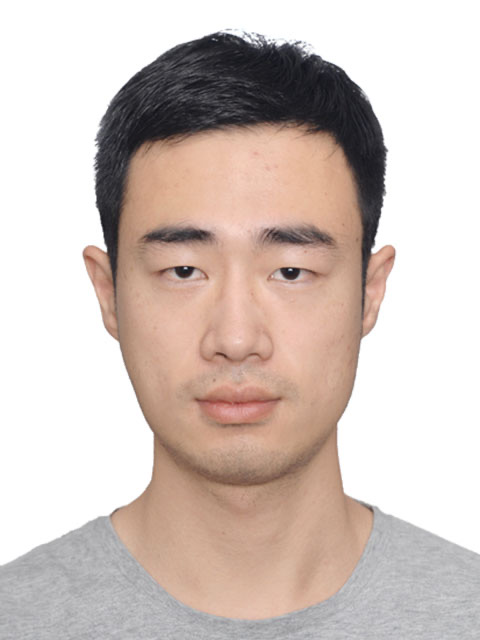}}]{Sheng Yang}
 is a PhD candidate at Department of Biomedical Engineering, Tsinghua University. Also a member of Guangzhi Wang's Lab, fortunately advised by Prof. Guangzhi Wang, with current research focus on Medical Robot, especially on Compute-Aided Image Guided Navigation System. He received the B.Sc degree and the M.Sc degree in Biomedical Engineering from Southeast University in 2016 and in 2019 respectively. His current interests include medical robot and medical image processing.
\end{IEEEbiography}

\begin{IEEEbiography}[{\includegraphics[width=1in,height=1.25in,clip,keepaspectratio]{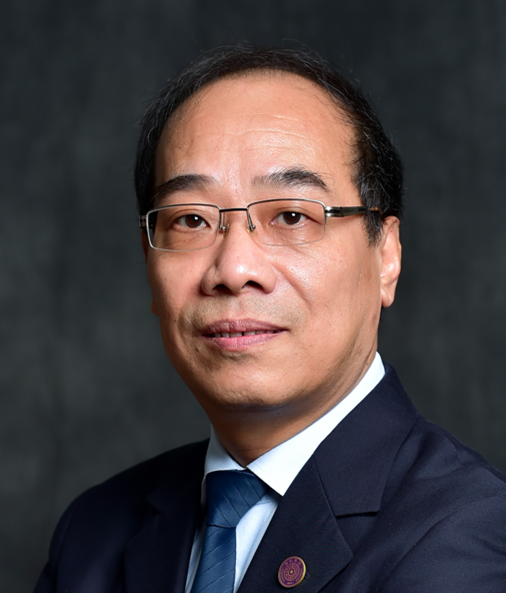}}]{Guangzhi Wang}, PhD, received his B.E. and M.S. Degrees in mechanical engineering in 1984 and 1987 respectively, and received his PhD Degree in Biomedical Engineering in 2003, all from Tsinghua University, Beijing, China. Since 1987, he has been with the Biomedical Engineering division at Tsinghua University. Since 2004, He is a tenured full Professor at Department of Biomedical Engineering, Tsinghua University. He is the author of hundreds of peer-reviewed scientific papers and is the inventor of 30 granted Chinese and PCT patents. Since 2015 Professor Guangzhi Wang was elected as the vice president of Chinese Society of Biomedical Engineering, and since 2018, he was elected as vice president of Chinese Association of Medical Imaging Technology. His current research interest includes biomedical image processing, image based surgical planning and computer aided surgery.
\end{IEEEbiography}

\begin{IEEEbiography}[{\includegraphics[width=1in,height=1.25in,clip,keepaspectratio]{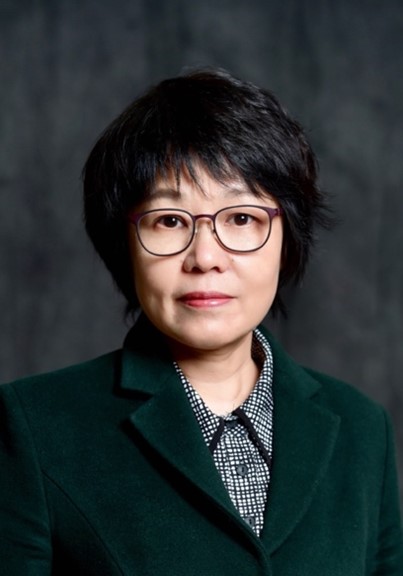}}]{Hui Ding} is a senior engineer and laboratory director in the Department of Biomedical Engineering, Tsinghua University. She was awarded the M.Sc degree in computer science from the University of science and technology of China. She has been engaged in medical image processing, computer aided minimally invasive treatment planning, human motion modeling and analysis for rehabilitation. At present, her research work includes multimode image guided minimally invasive orthopedic and neurosurgery, ultrasonic fusion imaging. She is the inventor of more than 10 Chinese and PCT patents.
\end{IEEEbiography}

\begin{IEEEbiography}[{\includegraphics[width=1in,height=1.25in,clip,keepaspectratio]{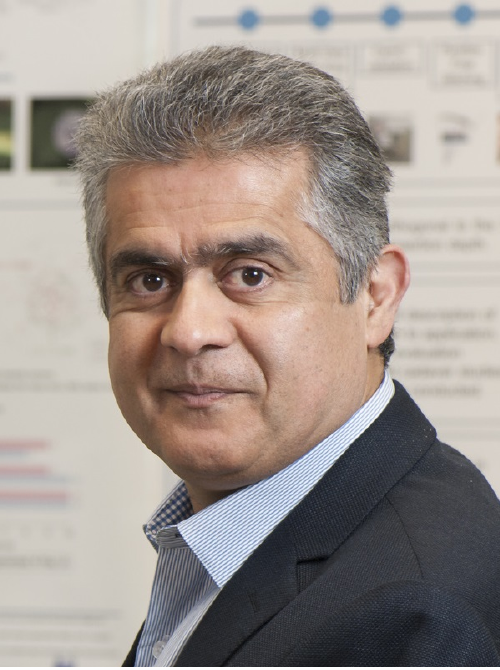}}]{Nassir Navab} is a Full Professor and Director of the Laboratory for Computer-Aided Medical Procedures (CAMP) at the Technical University of Munich (TUM) and Adjunct Professor at Laboratory for Computational Sensing and Robotics (LCSR) at Johns Hopkins University (JHU). He is a fellow of MICCAI society and received its prestigious Enduring Impact Award in 2021. He was also the recepient of 10 years lasting impact award of IEEE ISMAR in 2015, SMIT Society Technology award in 2010 and Siemens Inventor of the year award in 2001. He completed his PhD at INRIA and University of Paris XI, France, in January 1993, before enjoying two years of post-doctoral fellowship at MIT Media Laboratory and nine years of expeirence at Siemens Corporate Research in Princeton, USA.
He has acted as a member of the board of directors of the MICCAI Society, 2007-2012 and 2014-2017, and serves on the Steering committee of the IEEE Symposium on Mixed and Augmented Reality (ISMAR) and Information Processing in Computer-Assisted Interventions (IPCAI). He is the author of hundreds of peer-reviewed scientific papers, with more than 52,000 citations and an h-index of 101 as of April, 2022. He is the author of more than thirty awarded papers including 11 at MICCAI, 5 at IPCAI, 2 at IPMI and 3 at IEEE ISMAR. He is the inventor of 51 granted US patents and more than 60 International ones. His current research interests include medical augmented reality, computer-assisted surgery, medical robotics, computer vision and machine learning.
\end{IEEEbiography}

\begin{IEEEbiography}[{\includegraphics[width=1in,height=1.25in,clip,keepaspectratio]{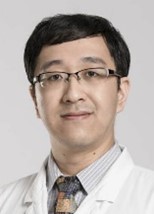}}]{Zhe Zhao}
is an attending surgeon of the Department of Orthopeadics, Beijing Tsinghua Changgung Hospital. Associate professor of School of Clinical Medicine, Tsinghua University. His research interest covers orthopaedic trauma, computer assisted surgery and orthopeadic implant development.
\end{IEEEbiography}

\begin{IEEEbiography}[{\includegraphics[width=1in,height=1.25in,clip,keepaspectratio]{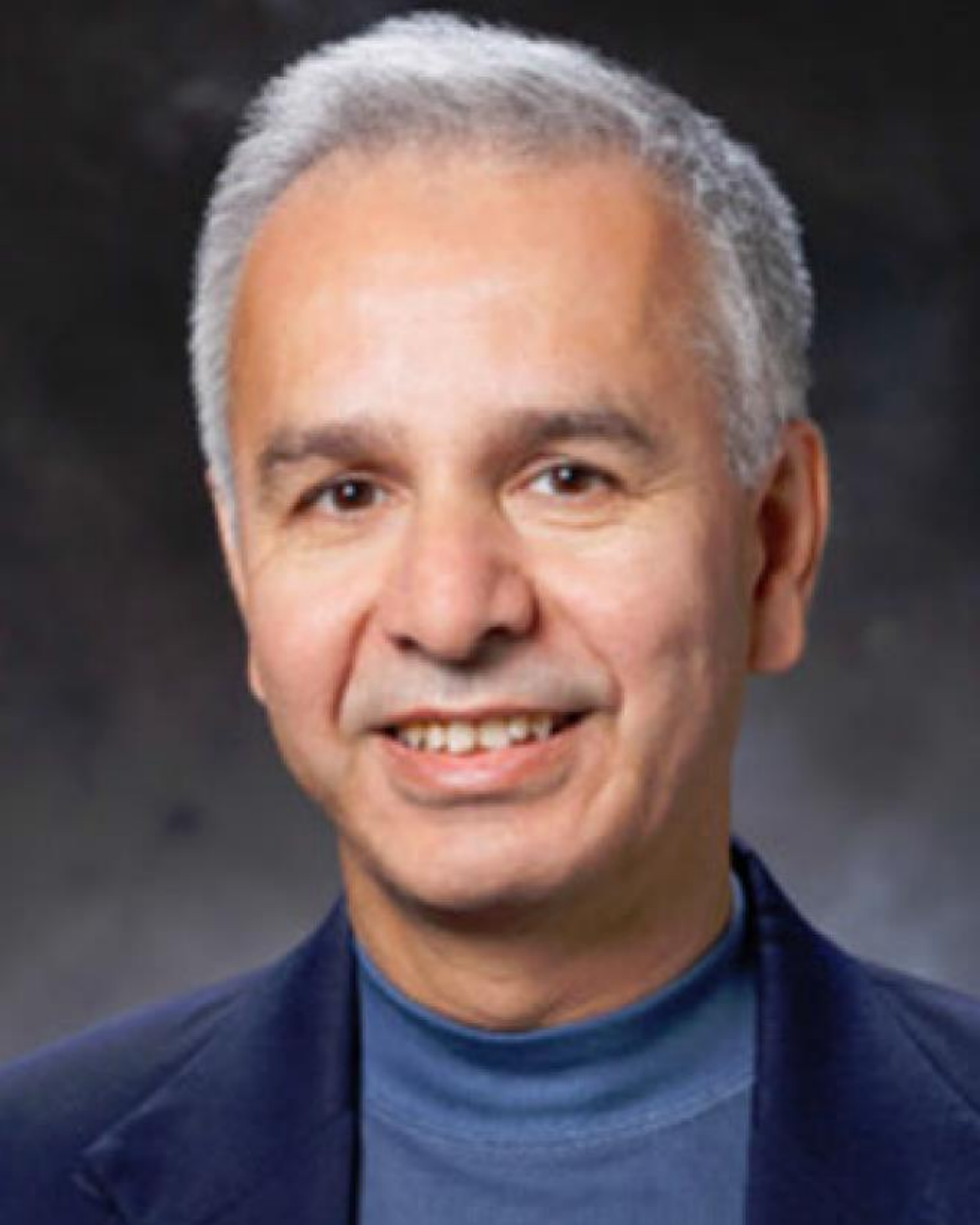}}]{Mehran Armand} received the Ph.D. degree in mechanical engineering and kinesiology from the University of Waterloo, Waterloo, ON, Canada, in 1998. He is currently a Professor of Orthopaedic Surgery, Mechanical Engineering, and Computer Science with Johns Hopkins University (JHU), Baltimore, MD, USA, and a Principal Scientist with the JHU Applied Physics Laboratory (JHU/APL). Prior to joining JHU/APL in 2000, he completed postdoctoral fellowships with the JHU Orthopaedic Surgery and Otolaryngology-Head and Neck Surgery. He currently directs the Laboratory for Biomechanical- and Image-Guided Surgical Systems, JHU Whiting School of Engineering. He also directs the AVICENNA Laboratory for advancing surgical technologies, Johns Hopkins Bayview Medical Center. His laboratory encompasses research in continuum manipulators, biomechanics, medical image analysis, and augmented reality for translation to clinical applications of integrated surgical systems in the areas of orthopaedic, ENT, and craniofacial reconstructive surgery.
\end{IEEEbiography}





\end{document}